# C2F-SemiCD: A Coarse-to-Fine Semi-Supervised Change Detection Method Based on Consistency Regularization in High-Resolution Remote-Sensing Images

Chengxi Han, *Student Member, IEEE*, Chen Wu, *Member, IEEE*, Meiqi Hu, *Student Member, IEEE*, Jiepan Li, *Student Member, IEEE*, Hongruixuan Chen, *Student Member, IEEE*

*Abstract*—A high-precision feature extraction model is crucial for change detection. In the past, many deep learning-based supervised change detection methods learned to recognize change feature patterns from a large number of labelled bi-temporal images, whereas labelling bi-temporal remote sensing images is very expensive and often time-consuming. Therefore, we propose a coarse-to-fine semi-supervised change detection method based on consistency regularization (C2F-SemiCD), which includes a coarse-to-fine change detection network with a multi-scale attention mechanism(C2FNet) and a semi-supervised update method. Among them, the C2FNet network "gradually" completes the extraction of change features from coarse-grained to fine-grained through multi-scale feature fusion, channel attention mechanism, spatial attention mechanism, global context module, feature refine module, initial aggregation module, and final aggregation module. The semi-supervised update method uses the mean teacher method. The parameters of the student model are updated to the parameters of the teacher Model by using the exponential moving average (EMA) method. Through extensive experiments on three datasets and meticulous ablation studies, including crossover experiments across datasets, we verify the significant effectiveness and efficiency of the proposed C2F-SemiCD method. The code will be open at: https://github.com/ChengxiHAN/C2F-SemiCD-and-C2FNet.

## I. INTRODUCTION

CHANGE detection (CD) is a remote sensing image analysis technique used to detect and identify changes in the surface or landscape that occur between different time points in the same area[1]. In change detection, remote sensing images from two or more time points are usually compared to identify differences or changes between images. These differences can be detected and quantified by pixel-level, object-level, or region-level analysis. High-resolution remote sensing images have a wide range of applications in change detection, including urban planning and land management[2],[3], environmental monitoring[4], resource management, natural disaster management[5], military intelligence analysis, and other fields[6],[7],[8],[9]. Change detection in high-resolution remote sensing images is important for monitoring surface changes, supporting decision-making, and providing timely information.

Machine learning especially deep learning has been applied in many fields of remote sensing, such as building extraction[10], [11],[12], hyperspectral remote sensing classification[13], [14], [15]and change detection[15], [16],[19],[20] and low-level task[21]. Now, many deep learning-based supervised methods have shown promising results in interpreting land-cover changes from bi-temporal VHR images. For example, there have been abundant CNN models (FC-EF[22], FC-Siam-Conc[22], FC-Siam-Diff[22]), attention mechanism-based models (HANet[23], HCGMNet[24], CGNet[25], MSF-Net[26], MFCN[26]), Transformer based models (BIT[28], Change Former[29], RSP-BIT[30]). Although deep learning-based methods have achieved good results, this result is heavily built on a large number of high-quality labelled samples. However, labelling samples of high-quality bi-temporal remote sensing images often need to compare the two images semantically and label the changed regions in the form of pixel level, which is very expensive and often time-consuming.

By contrast, in practical applications, it is easier to obtain a large number of unlabeled samples and a small number of labelled samples. It is worth studying how to make full use of the information provided by these small number of labelled samples and the large number of unlabeled samples to train the model. Therefore, the change detection method based on semi-supervised learning can overcome the above difficulties. With the help of only a small amount of labelled data, it combines unlabeled data with labelled data for analysis, so as to identify the change characteristics of the land surface more accurately. Some semi-supervised learning-based change detection methods include SemiCDNet[31], SemiCD[32], RCL[33], TCNet[34], [35], and some hot methods in the field of computer vision are fine-tuned for change detection tasks such as AdvNet[36], S4GAN[37], UniMatch-PSPNet[38], UniMatch-

This work was supported in part by the National Key Research and Development Program of China under Grant 2022YFB3903300 and 2022YFB3903405, and in part by the National Natural Science Foundation of China under Grant T2122014, 61971317, 62225113 and 42230108. (*Corresponding author: Chen Wu.*)

Chengxi Han, Chen Wu, Meiqi Hu and Jiepan Li are with the State Key Laboratory of Information Engineering in Surveying, Mapping and Remote Sensing, Wuhan University, Wuhan 430079, China (e-mail: chengxihan@whu.edu.cn; chen.wu@whu.edu.cn; meiqi.hu@whu.edu.cn; jiepanli@whu.edu.cn).

Hongruixuan Chen is with the Graduate School of Frontier Sciences, The University of Tokyo, Chiba 277-8561, Japan (e-mail: Qschrx@gmail.com).

DeepLabv3+[38], etc. The innovative ideas of these methods will be explained in detail in the experimental section. However, these semi-supervised CD methods still have room for improvement in distinguishing the changes of interest from the changes of irrelevant in feature extraction ability.

In order to effectively deal with the problem of data scarcity, that is the limited labelled data and abundant unlabeled data are fully utilized to improve the detection performance under the condition of the small model running time and inference time, we propose a more robust semi-supervised change detection method from coarse-grained to fine-grained, C2F-SemiCD, based on consistency regularization [39] [40]. In fact, our C2F-SemiCD includes two parts: C2FNet and mean teacher method [41] for semi-supervised learning to update parameters. Among them, C2FNet is designed to improve the encoding and decoding ability through multi-scale feature fusion, channel attention mechanism, spatial attention mechanism, global context module, feature refine module, initial aggregation module, and final aggregation module, for the purpose of improving the change feature extraction ability. The semi-supervised update method uses the mean teacher method to learn the change features, including the student model and the teacher model, both of which are based on C2FNet. The parameters of the student model are obtained by a small amount of training, and then the parameters of the teacher model are updated by the exponential moving average method. The teacher model generates pseudo-labels to guide the training of the student model, thereby improving the feature learning of unlabeled data. Since the pseudo-labels guide the subsequent unlabeled training, this method can achieve a semi-supervised effect. It can be seen from Fig. 1 that our proposed method has a significant improvement effect compared with the proposed supervised method and solves the missed detection phenomenon (less blue area) and false detection phenomenon (less red area) well in visual features. Even when the proportion of labels in the training data is only 30%, the results are close to the results of the fully supervised method when the proportion of labels is 100%, which further proves the effectiveness of our proposed semi-supervised method.

Therefore, the main contributions of this paper are:

(1) A novel semi-supervised change detection approach, denoted as C2F-SemiCD, is introduced in this study. The method employs consistency regularization to augment the feature learning capability for both labelled and unlabeled data, which is achieved by minimizing the disparity between predictions made by the student model and a corresponding teacher model. Significantly, C2F-SemiCD exhibits notable advantages in terms of both training and inference time efficiency.

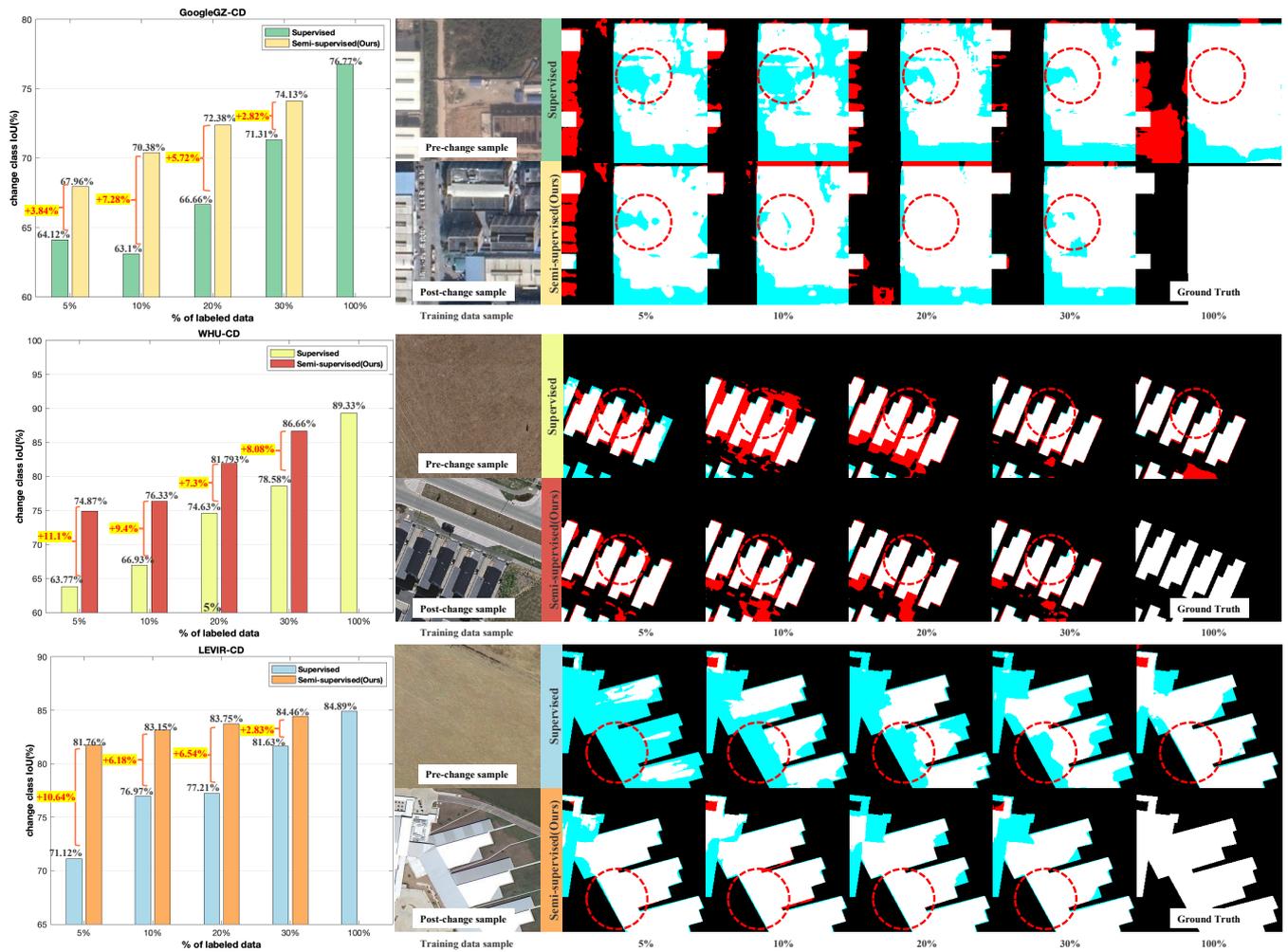

**Fig. 1.** It is verified on three datasets that the proposed semi-supervised method can effectively obtain the quantitative statistics and visualization effects of a large number of unlabeled images. TP (white), TN (black), FP (red), and FN (blue).

(2) A coarse-to-fine change detection network with a multi-scale attention mechanism (C2FNet) is proposed, which gradually completes the extraction of change features from coarse to fine granularity through multi-scale feature fusion, channel attention mechanism, spatial attention mechanism, global context module, feature refine module, initial aggregation module, and final aggregation module. Remarkably, the feature extraction proficiency outperforms other semi-supervised methods in a subset of labelled samples.

(3) Conducting extensive experiments and detailed ablation studies across four scales (5%, 10%, 20%, 30%) on three prominent binary change detection datasets, along with cross-dataset experiments, solidly establishes the remarkable effectiveness and efficiency of the proposed method. Moreover, we are delighted to contribute significantly to the domain of change detection research by open-sourcing the source code, thereby promoting transparency and fostering collaborative efforts within the scientific community.

The rest of this paper is organized as follows. In Section II, the structure of the proposed C2F-SemiCD and C2FNet are illustrated in detail. Section III reports the experimental results and ablation study. Finally, Section IV concludes the article.

## II. METHODOLOGY

In this section, we introduce the proposed C2F-SemiCD and then present the model details. Fig. 2. shows the illustration of our proposed C2F-SemiCD.

### A. *C2F-SemiCD Framework*

Semi-supervised learning can be categorized into three main groups: pseudo-labeling, methods based on generative adversarial networks, and consistency learning techniques [31]. Consistency learning methods enhance both the robustness of models and the quality of segmentation results by reducing discrepancies in predictions across various perturbed versions of the same unlabeled sample. Therefore, we design a coarse-to-fine semi-supervised change detection method based on consistency regularization, as shown in Fig. 2, which aims to solve how to learn image features from large amounts of data without labels.

In semi-supervised learning tasks, for unlabeled data, we use the mean teacher[41] method to learn the features. The main idea of the model is to switch between the roles of the student model and the teacher model, specifically, the teacher model is the average of successive student models. When the backbone network is used as the teacher model, it mainly produces the goal of the student model. When the backbone network is used as the student model, the objectives generated by the teacher model are used for learning. And, the parameters of the teacher model are obtained by the weighted average of the parameters of the student model trained in the previous iterations. The target label of the unlabeled data comes from the prediction result of the teacher model, that is, the pseudo-label. Because the pseudo-labels generated by the teacher model can be used as prior information in the later training to guide the student model to learn a large number of features without labelled data, the effect of semi-supervised learning can be achieved. If C2FNet is trained as a separate network, the whole training process is the supervised phase, while C2F-SemiCD is the training of the first few epochs with labels and the following epochs without labels, so it is called the unsupervised phase. The quality of the generated pseudo-labels plays a crucial role in the performance of the final model, so in the process of supervised training, we propose a robust coarse-to-fine change detection network with a multi-scale attention mechanism (C2FNet), as shown in Fig. 3 and Fig. 4, which is described in detail below. The number of supervised training is 5, which will also be verified in the ablation experiment. Instead of sharing weights with the student model, the teacher model updates the parameters of the teacher model via the exponential moving average (EMA) of the student model. By defining the consistency loss $J$ as the expected distance between the predictions of the student model (weights $\theta$ and perturbations $\eta$) and the predictions of the teacher model

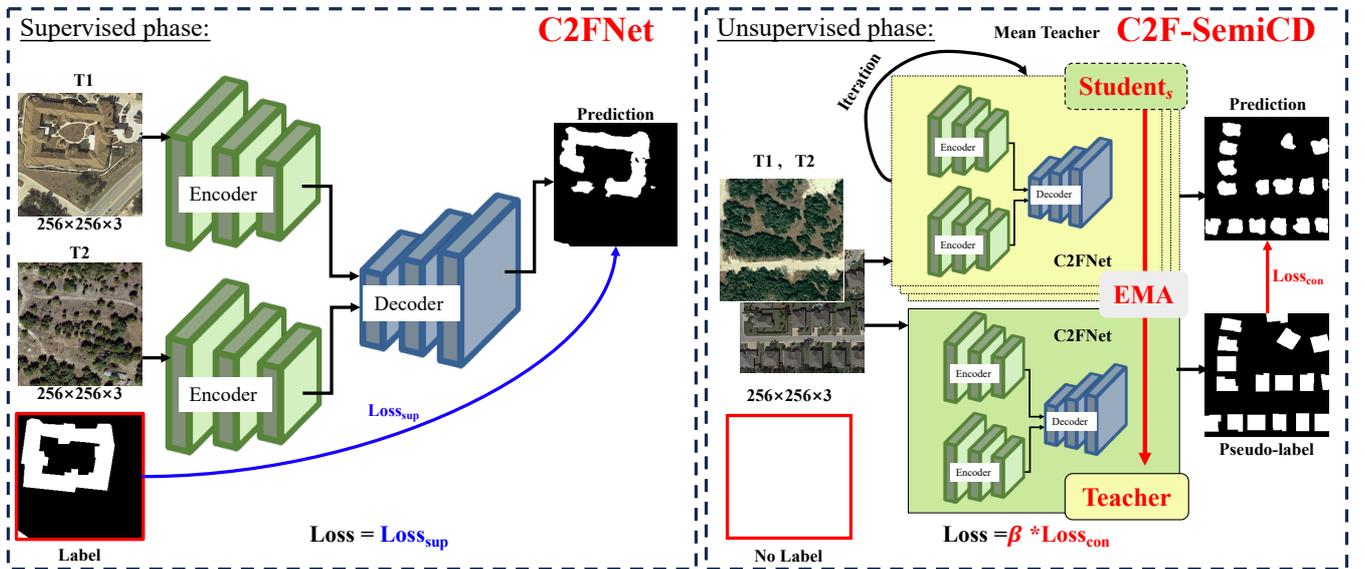

**Fig. 2.** Overall framework of the proposed C2F-SemiCD. C2F-SemiCD includes C2FNet and the mean teacher method for semi-supervised learning to update parameters.

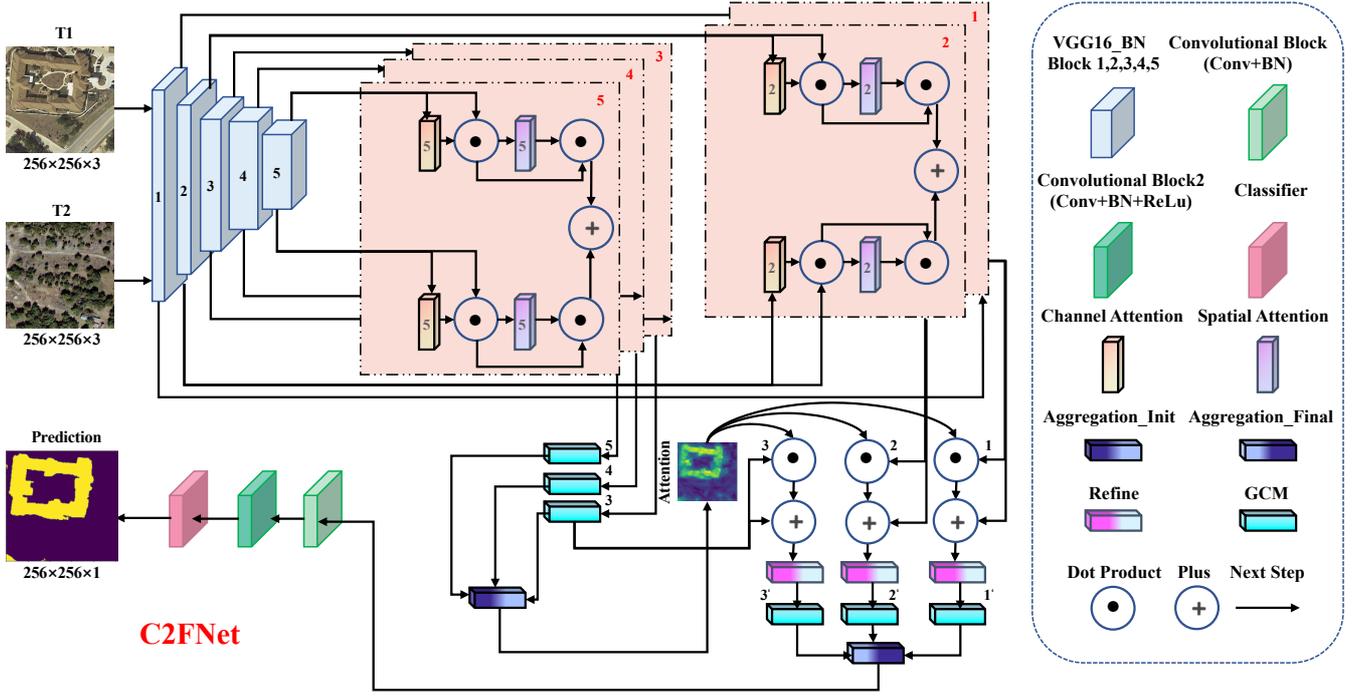

**Fig. 3.** Illustration of a coarse-to-fine change detection network (C2FNet).

(weights $\theta'$ and perturbations $\eta'$):

$$J(\theta) = \mathbb{E}_{x,\eta',\eta}[||f(x,\theta',\eta') - f(x,\theta,\eta)||^2] \quad (1)$$

We do this by defining $\theta'_t$ as the EMA with continuous $\theta$ weights for training phase $t$:

$$\theta'_t = \alpha\theta'_{t-1} + (1-\alpha)\theta_s \quad (2)$$

where $\alpha$ is a super smooth coefficient parameter, the experiment of $\alpha$ is 0.99, $\theta'_t$ is the new parameter of the teacher model, $\theta'_{t-1}$ is the old parameter of the teacher model, $\theta_s$ is the parameter of the student model. We can approximate the consistency loss function $J$ by sampling the perturbations $\eta$, $\eta'$ using stochastic gradient descent at each training step.

*B. A Coarse-to-Fine Change Detection Network (C2FNet*

1) The overall framework of the proposed C2FNet from Fig. 3: Firstly, in the Encoder part, we use the VGG-16[42] network as the backbone to extract the change features of the dual-temporal image from coarse to fine. We use five VGG16_BN blocks, which represent the operation of VGG16_BN from 0-5,5-12,12-22, 22-32, and 32-42 layers. Then, in the decoder part, the change features are extracted from coarse-grained to fine-grained from Fig. 5. by means of multi-scale feature fusion, channel attention mechanism[43], spatial attention mechanism[43], global context module, feature refine module, initial aggregation module, and final aggregation module. The change process from coarse-grained to fine-grained will also be shown in detail in the visual features in Fig. 5.

2) Channel Attention: Attention mechanisms are methods that can focus attention on important areas of the image and discard irrelevant ones in the field of computer vision. Because they can greatly improve performance, they are widely used in deep learning models. The channel attention mechanism mainly tells the network "what to focus on", which successively includes adaptive max pooling operation, 2D convolution operation (output dimension divided by 16), the activation function of ReLU, 2D convolution operation (input dimension divided by 16), and Sigmoid activation function from Fig. 4.

3) Spatial Attention: The spatial attention mechanism mainly solves the problem of telling the network "where to focus", which successively includes the row-wise tensor maximum, 1D convolution operation, and sigmoid activation function from Fig. 4. It is worth noting that the 2D convolution operation operates horizontally first and then vertically, while the 1D convolution operation can only operate vertically. Through the channel attention mechanism and spatial attention mechanism, the calculation amount can be greatly reduced.

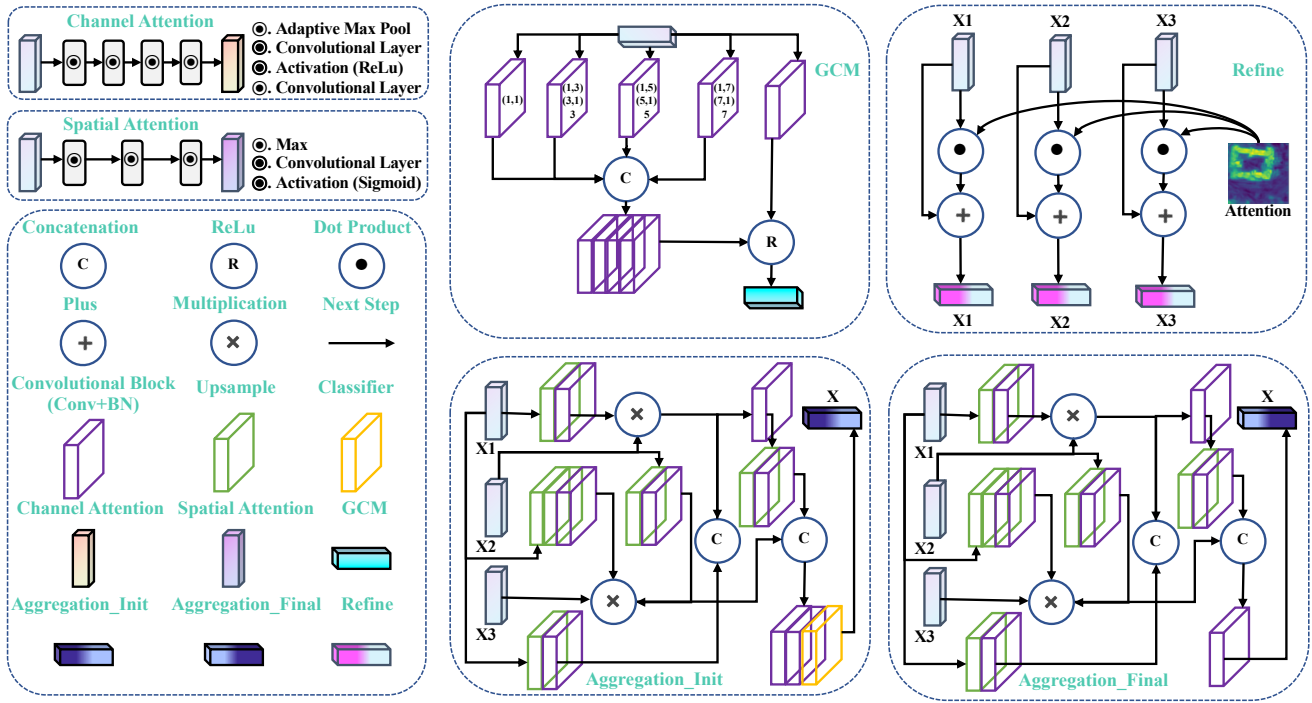

**Fig. 4.** Detailed explanation of the relevant modules involved in C2FNet.

4) Global Contextual Module(GCM): Firstly, a basic 2D convolution operation module is constructed, including 2D convolution operation and 2D batch normalization layer. The global context module consists of four ordered containers from Fig. 4. The first container contains a basic 2D convolution operation module, and the second to fourth containers contain four basic 2D convolution operation modules. Each basic 2D convolution operation module contains atrous convolutions with different atrous rates to ensure multi-scale context information of features. We concatenate four tensors in a horizontal manner and put the concatenation results into the activation function of ReLU along with the input values from a basic 2D convolution operation. Multi-scale context information can be obtained through this global context module.

5) Refine Module: The feature refine module includes three inputs and three outputs from Fig. 4, of which three inputs are related to the attention map, the high-level features are upsampled and the low-level features are dot multiplied to eliminate the differences between the features, and the first stage of change decoding is completed in turn. After 4 times bilinear upsampling, the Attention map is dot multiplied by X1, and the result of the operation is added to the original input X1 to obtain the final X1. Attention The attention map is bilinear upsampled by a factor of 2 and then dot multiplied with X2. The result is added to the original input X2 to obtain the final X2. The Attention map is dot multiplied by X3, and the result is added to the original input X3 to get the final X3.

6) Aggregation_Init: The initial aggregation module includes the input of three feature stages and output from Fig. 4. The three-layer feature map is adjusted based on the change decoding map (highlighting the positive sample), and then more refined decoding is performed by the decoding method of the feature improvement module, so as to improve the accuracy of the change information. The detailed operation process is that the deep input feature X1 is subjected to 2 times bilinear upsampling and then the basic 2D convolution operation module is performed, and the operation result is multiplied with the deeper input feature X2 to obtain the result 1. The result of the deeper input feature X2 is multiplied by the result of the basic 2D convolution operation module after twice bilinear upsampling, and the result of the deep input feature X1 is multiplied by the basic 2D convolution operation module. The result of the operation is multiplied by the basic 2D convolution operation module and the middle deep feature X3 to obtain result 2. The result of the deep input feature X1 after double bilinear upsampling is horizontally concatenated with result 1, and the concatenation result is obtained after the basic 2D convolution operation module with twice the input and output dimensions. The basic 2D convolution operation module with two times input and output dimensions was performed on result 3 after double bilinear up-sampling, and the operation results were horizontally spliced with result 2. The splicing results were performed on the basic 2D convolution operation module with three times input and output dimensions to obtain result 4. Result 4 is followed by a basic 2D convolution operation with three times the input and output dimensions. Result 5 Performing a normal 3x 2D convolution operation yields the final decoded value of the initial aggregation module.

7) Aggregation_Final: The function of the final aggregation module is the same as that of the initial aggregation module, and the operation process is the same from Fig. 4. The only difference is that the final operation result is result 4 instead of the last two steps. It is worth noting that these modules are represented by the last block of the operation to represent the whole operation process of this module. The initial aggregation module operates in the same way as the final aggregation module, so we represent it separately by alternating colour blocks.

To verify the effectiveness of GCM, Refine, Aggregation_Init, and Aggregation_Final modules, we visualize the decoding stage of feature extraction against the network schematic of C2FNet, as shown in Fig. 5. We use the three numbers in parentheses to represent the length, width, and thickness of the feature tensor, respectively, and take one of the dimensions to show the features at this stage. Firstly, after the feature maps A, B and C passed through the GCM module, the features of the buildings were significantly improved. Secondly, the Attention maps obtained by A1, B1 and C1 after the Aggregation_Init module can fuse the multi-scale features well, and distinguish the buildings from the background to a certain extent. Thirdly, D, E, C1, and the Attention map after the Refine module, the change features of buildings are obviously distinguished from the background, but the outline of the buildings is not clear. Fourth, after the GCM and Aggregation_Final modules of C2, D2 and E2 the foreground is highlighted (the colour of the change information is obvious) and the background is suppressed (the non-change information is suppressed obviously). In general, change features are extracted from coarse intensity to fine granularity through these modules.

## C. Loss Function

Because the change detection task can be seen as a special binary classification problem, we used a binary cross-entropy loss function, combining Sigmoid and BCELoss.

$$Loss_{sup} = \{l_1, \dots, l_N\} \tag{3}$$

$$l_n = -[y_n \cdot log(\sigma(X_n)) + (1 - Y_n) \cdot log(1 - \sigma(X_n))] \tag{4}$$

$$\sigma(X) = \frac{1}{1+exp(-x)} \tag{5}$$

$\sigma(X_n)$ is a Sigmoid function that maps x to the interval (0,1). In the supervised training part, the input is the ground truth, and the supervised loss function is:

$$Loss = Loss_{sup} \tag{6}$$

In the semi-supervised training part, the input is the result of the supervised training, and the semi-supervised loss function:

$$Loss_{con} = -[\hat{y}_t \cdot log(\sigma(X_n)) + (1 - \hat{y}_t) \cdot log(1 - \sigma(X_n))] \tag{7}$$

$$\hat{y}_t = \frac{1}{1+exp(-\widehat{x_t})} \tag{8}$$

$$Loss = Loss_{sup} + \beta \cdot Loss_{con} \tag{9}$$

here, we used 0.2 for $\beta$, which will be verified in the ablation study.

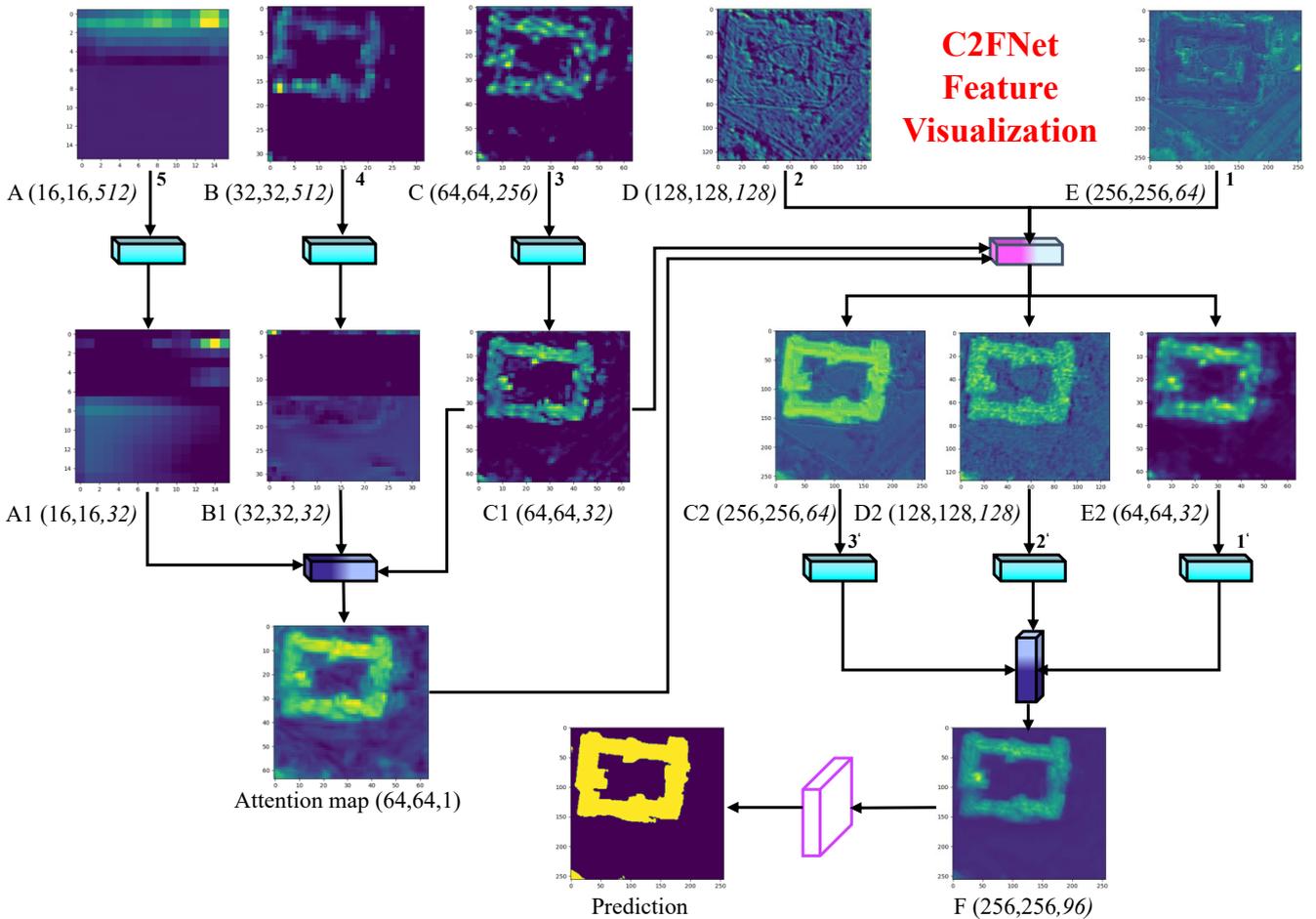

**Fig. 5.** Visualization process of feature extraction in the decoding stage of C2FNet.

TABLE I
NUMBER OF TRAIN (LABELLED AND UNLABELED), VAL, AND TEST IMAGES IN THE DATASET FOR SEMI-SUPERVISED EXPERIMENTS.

| Labelled Ration | Train (Image Number) | | | | | |
|---|---|---|---|---|---|---|
| | GoogleGZ-CD[31] | | WHU-CD[44] | | LEVIR-CD [45] | |
| | Labelled | Unlabelled | Labelled | Unlabelled | Labelled | Unlabelled |
| 5% | 37 | 703 | 298 | 5651 | 357 | 6765 |
| 10% | 74 | 666 | 595 | 5354 | 713 | 6409 |
| 20% | 148 | 592 | 1190 | 4759 | 1425 | 5697 |
| 30% | 222 | 518 | 1785 | 4164 | 2137 | 4985 |
| Val / Test Image Number | 116 | 300 | 744 | 745 | 1025 | 2049 |

## III. EXPERIMENT

In this section, we verify the effectiveness of our proposed method through extensive experiments, mainly including the introduction of the experimental environment, comparison methods, experimental data, evaluation indicators, and ablation experiments.

### A. Experimental setup

*GooglGZ-CD*[31]: A large-scale Very High-Resolution (VHR) multispectral satellite image dataset openly released by Wuhan University. It encompasses 20 pairs of seasonal high-resolution images captured in the suburban areas of Guangzhou, China, spanning from 2006 to 2019. These images vary in dimensions from 1006×1168 pixels to 4936×5224 pixels, with a spatial resolution of 0.55 meters. The dataset encompasses diverse image change types such as water bodies, roads, farmland, bare land, forests, buildings, and vessels. For GPU training convenience, the image pairs have been cropped into non-overlapping blocks of size 256×256 pixels.

*WHU-CD*[44]: A large-scale Very High-Resolution (VHR) dataset for building change detection, publicly provided by Wuhan University. It comprises a high-resolution image pair of dimensions 32507 × 15345, captured in the Christchurch region of New Zealand in 2011 and subsequent years, featuring a spatial resolution of 0.2 meters. Due to significant reconstruction following an earthquake, the primary change information primarily revolves around buildings. To account for GPU memory limitations and ensure a fair comparison with other algorithms, the default image pairs were directly cropped into non-overlapping blocks of size 256 × 256.

*LEVIR-CD*[45]: A substantial Very High-Resolution (VHR) building change detection dataset made available by Beihang University. It comprises 637 pairs of high-resolution image blocks, captured from 2002 to 2018, with image dimensions of 1024 × 1024 pixels and a spatial resolution of 0.5 meters. The dataset predominantly focuses on building-related changes and encompasses various types of structures, including villa residences, high-rise apartments, small garages, and large warehouses. To accommodate GPU memory limitations and ensure equitable comparisons with other algorithms, the default image pairs were directly cropped into non-overlapping blocks of size 256 × 256.

In the semi-supervised experiments, the datasets were constructed through random sampling. Specifically, 5%, 10%, 20%, and 30% of the training data were randomly chosen as labelled data, while the remaining data were treated as unlabeled data. For a comprehensive breakdown, refer to Table I.

*Implementation details*. To validate the effectiveness of the proposed method, we reproduced the model on PyTorch and conducted training and testing on a single NVIDIA RTX 3090 GPU. For the specific parameter configuration of the model, we utilized the AdamW optimizer to minimize the loss function, with a weight decay coefficient of 0.0025 and a learning rate of 5e-4. Due to GPU limitations, extensive experimentation led us to set the batch size and epoch number to 16 and 100, respectively, to achieve model convergence. During each training epoch, testing was performed on the validation set. During the model saving process, we compared F1 scores and IoU scores, and ultimately saved the model with the highest simultaneous F1 and IoU scores as the best-performing model.

*Evaluation Metrics*. To verify the performance of the proposed method, F1-score (F1), Precision (Pre.), Recall (Rec.), Overall Accuracy (OA), Kappa Coefficient (KC), and Intersection over the Union (IoU) are used for quantitative evaluation. They can be defined as follows:

$$F1 = \frac{2}{Pre.^{-1} + Rec.^{-1}} \quad (10)$$

$$Pre. = \frac{TP}{TP+FP} \quad (11)$$

$$Rec. = \frac{TP}{TP+FN} \quad (12)$$

$$OA = \frac{TP+TN}{TP+TN+FN+FP} \quad (13)$$

$$KC = \frac{OA-PRE}{1-PRE} \quad (14)$$

$$IoU = \frac{TP}{TP+FN+FP} \quad (15)$$

$$PRE = \frac{(TP+FN)\times(TP+FP)}{(TP+TN+FP+FN)^2} + \frac{(TN+FP)\times(TN+FN)}{(TP+TN+FP+FN)^2} \quad (16)$$

where TP denotes the number of true positives, TN denotes the number of true negatives, FP denotes the number of false positives, and FN denotes the number of false negatives. The "PRE" is the sum of the "ground truth and the product of the predicted result" corresponding to all categories, divided by the "average of the total sample data". It's important to highlight that higher values of F1, IoU, and KC indicate effective Change Detection performance.

### B. Comparison with state-of-the-art methods

1) **Supervised-Only: C2FNet.**
2) AdvNet[36]: A semi-supervised semantic segmentation method utilizing adversarial networks in the field of computer vision, where the discriminator enhances semantic segmentation accuracy through combined adversarial loss and standard cross-entropy loss, adapted suitably for change detection tasks by appropriate modifications.

3) S4GAN[37]: A semi-supervised framework in the field of computer vision, utilizing a generator network from Generative Adversarial Networks (GANs) to provide additional training samples to a multi-class classifier. The primary concept is to approximate the feature space of real samples by incorporating a substantial amount of synthetic visual data, thereby enhancing multi-class pixel classification tasks, adapted suitably for change detection tasks by appropriate modifications.

4) SemiCDNet[31]: A semi-supervised convolutional change detection network based on Generative Adversarial Networks (GANs), where inputs with both labelled and unlabeled data are fed into the segmentation network to generate initial predictions and entropy maps. Simultaneously, two discriminators are employed to enhance the feature distribution consistency between segmentation maps and entropy maps of labelled and unlabeled data. Throughout the training process, the generator continually utilizes unlabeled information for regularization, thereby enhancing its generalization capability.

5) SemiCD[32]: A semi-supervised CD model enforces consistency in the output change probability maps of provided unlabeled dual-temporal image pairs, even under slight random perturbations applied to the deep feature difference maps. These maps are derived by subtracting their latent feature representations. Subsequently, unsupervised CD loss and supervised Cross-Entropy (CE) loss are computed based on this principle.

6) RCL[33]: A reliable contrastive learning technique tailored for semi-supervised Change Detection (CD) in remote sensing imagery. It introduces a contrastive loss rooted in change regions to bolster the model's feature extraction from changed objects. By leveraging the ambiguity of unlabeled data, it effectively identifies dependable pseudo-labels for model training. Through the fusion of these tactics, it optimally taps into the latent distinctive features inherent in unlabeled data.

TABLE II

QUANTITATIVE ACCURACY RESULTS OF SOTA COMPARISON METHODS WITH DIFFERENT LABELLED RATIOS ON THE GOOGLEGZ-CD DATASET. RED HIGHLIGHTS THE BEST VALUES.

| Method | Labelled Ration | | | | | | | | | | | |
|---|---|---|---|---|---|---|---|---|---|---|---|---|
| | 5% | | | | | | 10% | | | | | |
| | F1 | Pre. | Rec. | OA | KC | IoU | F1 | Pre. | Rec. | OA | KC | IoU |
| Super. Only (C2FNet) | 78.14 | 77.49 | 78.80 | 89.16 | 70.93 | 64.12 | 77.37 | 82.65 | 72.73 | 89.55 | 70.61 | 63.10 |
| AdvNet | 59.34 | 67.64 | 52.86 | 82.20 | 48.16 | 42.19 | 68.14 | 76.87 | 61.18 | 85.94 | 59.26 | 51.67 |
| S4GAN | 64.08 | 70.30 | 58.87 | 83.78 | 53.71 | 47.14 | 69.63 | 77.41 | 63.27 | 86.43 | 61.01 | 53.41 |
| SemiCDNet | 59.57 | 63.85 | 55.83 | 81.37 | 47.54 | 42.42 | 64.33 | 73.43 | 57.24 | 84.40 | 54.55 | 47.42 |
| SemiCD | 60.44 | 70.33 | 52.99 | 82.95 | 49.85 | 43.31 | 68.12 | 75.45 | 62.09 | 85.72 | 59.03 | 51.66 |
| RCL | 71.43 | 74.90 | 68.27 | 86.58 | 62.69 | 55.56 | 77.51 | 76.78 | 78.25 | 88.84 | 70.09 | 63.28 |
| TCNet | 74.75 | 78.97 | 70.95 | 88.22 | 67.09 | 59.68 | 78.68 | 75.79 | 81.79 | 89.10 | 71.37 | 64.85 |
| UniMatch -PSPNet | 42.36 | 93.43 | 27.39 | 81.68 | 35.13 | 26.87 | 56.92 | 74.27 | 46.14 | 82.83 | 46.92 | 39.78 |
| UniMatch -DeepLabv3+ | 52.50 | 61.86 | 45.60 | 79.72 | 39.98 | 35.59 | 63.93 | 81.79 | 52.47 | 85.45 | 55.35 | 46.98 |
| **C2F-SemiCD** | **80.93** | **81.05** | **80.80** | **90.64** | **74.72** | **67.96** | **82.61** | 79.04 | **86.53** | **91.05** | **76.60** | **70.38** |
| Method | Labelled Ration | | | | | | | | | | | |
| | 20% | | | | | | 30% | | | | | |
| | F1 | Pre. | Rec. | OA | KC | IoU | F1 | Pre. | Rec. | OA | KC | IoU |
| Super. Only (C2FNet) | 79.99 | 89.51 | 72.31 | 91.11 | 74.36 | 66.66 | 83.26 | 87.64 | 79.29 | 92.16 | 78.15 | 71.31 |
| AdvNet | 69.90 | 85.54 | 59.09 | 87.49 | 62.33 | 53.72 | 70.25 | 83.27 | 60.75 | 87.35 | 62.47 | 54.14 |
| S4GAN | 71.16 | 87.44 | 59.99 | 88.05 | 63.95 | 55.23 | 72.71 | 82.78 | 64.83 | 88.04 | 65.20 | 57.12 |
| SemiCDNet | 70.78 | 87.54 | 59.40 | 87.94 | 63.53 | 54.77 | 72.33 | 85.13 | 62.87 | 88.18 | 65.02 | 56.65 |
| SemiCD | 71.75 | 88.25 | 60.45 | 88.30 | 64.70 | 55.95 | 72.97 | 80.90 | 66.46 | 87.90 | 65.27 | 57.44 |
| RCL | 79.44 | 78.82 | 80.07 | 89.81 | 72.67 | 65.89 | 80.84 | 83.30 | 78.53 | 90.85 | 74.84 | 67.85 |
| TCNet | 82.43 | 85.46 | 79.62 | 91.66 | 76.98 | 70.12 | 83.13 | 84.34 | 81.96 | 91.82 | 77.74 | 71.13 |
| UniMatch -PSPNet | 61.98 | 81.34 | 50.06 | 84.90 | 53.22 | 44.90 | 52.98 | 89.07 | 37.71 | 83.55 | 44.93 | 36.04 |
| UniMatch -DeepLabv3+ | 67.97 | 79.17 | 59.55 | 86.21 | 59.40 | 51.48 | 57.59 | 90.77 | 42.17 | 84.73 | 49.75 | 40.44 |
| **C2F-SemiCD** | **83.98** | 80.25 | **88.07** | **91.74** | **78.43** | **72.38** | **85.14** | 85.16 | **85.12** | **92.70** | **80.30** | **74.13** |

| Super. only | (C2FNet): 100% → | | | | 86.86 | 85.46 | 88.31 | 93.43 | 82.48 | 76.77 |

TABLE III
QUANTITATIVE ACCURACY RESULTS OF SOTA COMPARISON METHODS WITH DIFFERENT LABELLED RATIOS ON THE WHU-CD DATASET. RED HIGHLIGHTS THE BEST VALUES.

| Method | Labelled Ration | | | | | | | | | | | |
| --- | --- | --- | --- | --- | --- | --- | --- | --- | --- | --- | --- | --- |
| | 5% | | | | | | 10% | | | | | |
| | F1 | Pre. | Rec. | OA | KC | IoU | F1 | Pre. | Rec. | OA | KC | IoU |
| Super. only (C2FNet) | 77.88 | 85.80 | 71.29 | 98.39 | 77.05 | 63.77 | 80.19 | 81.72 | 78.72 | 98.46 | 79.39 | 66.93 |
| AdvNet | 71.10 | 78.76 | 64.80 | 97.91 | 70.03 | 55.16 | 77.96 | 79.97 | 76.06 | 98.29 | 77.08 | 63.89 |
| S4GAN | 73.39 | 81.56 | 66.70 | 98.08 | 72.40 | 57.96 | 76.53 | 77.85 | 75.25 | 98.17 | 75.57 | 61.98 |
| SemiCDNet | 70.07 | 79.97 | 62.36 | 97.89 | 68.99 | 53.93 | 77.69 | 79.62 | 75.85 | 98.27 | 76.80 | 63.52 |
| SemiCD | 78.76 | 86.22 | 72.49 | 98.45 | 77.96 | 64.96 | 81.49 | 82.51 | 80.49 | 98.55 | 80.73 | 68.76 |
| RCL | 71.56 | 74.63 | 68.73 | 97.83 | 70.43 | 55.71 | 77.69 | 78.24 | 77.15 | 98.24 | 76.78 | 63.52 |
| TCNet | 85.36 | 87.05 | 83.73 | 98.86 | 84.76 | 74.45 | 85.98 | 90.33 | 82.03 | 98.94 | 85.43 | 75.41 |
| UniMatch-PSPNet | 87.23 | 93.50 | 81.74 | 99.05 | 86.74 | 77.35 | 88.49 | 89.89 | 87.13 | 98.97 | 87.95 | 79.36 |
| UniMatch-DeepLabv3+ | 88.35 | 92.56 | 84.51 | 99.12 | 87.90 | 79.14 | 88.58 | 89.29 | 87.88 | 99.10 | 88.11 | 79.50 |
| C2F-SemiCD | 85.63 | 86.51 | 84.77 | 98.87 | 85.04 | 74.87 | 86.58 | 87.35 | 85.81 | 98.94 | 86.03 | 76.33 |
| Method | Labelled Ration | | | | | | | | | | | |
| | 20% | | | | | | 30% | | | | | |
| | F1 | Pre. | Rec. | OA | KC | IoU | F1 | Pre. | Rec. | OA | KC | IoU |
| Super. only (C2FNet) | 85.47 | 89.35 | 81.92 | 98.90 | 84.90 | 74.63 | 88.00 | 93.57 | 83.06 | 99.10 | 87.54 | 78.58 |
| AdvNet | 82.22 | 78.40 | 86.44 | 98.52 | 81.45 | 69.81 | 85.85 | 87.55 | 84.21 | 98.90 | 85.28 | 75.21 |
| S4GAN | 79.75 | 76.72 | 83.04 | 98.33 | 78.88 | 66.32 | 82.22 | 78.40 | 86.44 | 98.52 | 81.45 | 69.81 |
| SemiCDNet | 80.73 | 75.05 | 87.32 | 98.35 | 79.87 | 67.68 | 84.65 | 84.61 | 84.69 | 98.78 | 84.02 | 73.39 |
| SemiCD | 88.74 | 89.88 | 87.63 | 99.12 | 88.28 | 79.75 | 88.16 | 89.30 | 87.05 | 99.07 | 87.68 | 78.83 |
| RCL | 83.61 | 80.52 | 86.95 | 98.65 | 82.91 | 71.84 | 86.08 | 85.12 | 87.05 | 98.88 | 85.49 | 75.56 |
| TCNet | 88.92 | 94.56 | 83.91 | 99.17 | 88.49 | 80.05 | 89.34 | 92.49 | 86.39 | 99.18 | 88.91 | 80.73 |
| UniMatch-PSPNet | 89.46 | 86.71 | 92.39 | 99.14 | 89.01 | 80.93 | 89.67 | 88.31 | 91.07 | 99.17 | 89.24 | 81.28 |
| UniMatch-DeepLabv3+ | 89.26 | 86.27 | 92.47 | 99.12 | 88.80 | 80.60 | 90.47 | 89.06 | 91.93 | 99.23 | 90.07 | 82.60 |
| C2F-SemiCD | 90.07 | 91.83 | 88.36 | 99.23 | 89.66 | 81.93 | 92.85 | 93.84 | 91.88 | 99.44 | 92.56 | 86.66 |
| Super. only | (C2FNet): 100% → | | | | | | 94.36 | 96.57 | 92.26 | 99.56 | 94.14 | 89.33 |

7) TCNet[34], [35]: TCNet is a time-consistency network developed for semi-supervised change detection. Its focus lies in reinforcing the alignment of predictions stemming from various input sequences to effectively capture the distribution of unlabeled data. When dealing with labelled samples, two segmentation networks of identical architecture are trained using two distinct input sequences. For unlabeled samples, these two segmentation networks are individually utilized for forward prediction, resulting in two change detection outcomes. Subsequently, the generation of supervised signals involves minimizing the disparity between the two predicted outcomes.
8) UniMatch-PSPNet[38]: A semi-supervised approach in the field of computer vision that expands the perturbation space through auxiliary feature perturbation streams. Simultaneously, a dual-stream perturbation method is employed, enabling two robust views to be jointly guided by a shared weak view. The foundational framework is based on PSPNet, adapted accordingly for change detection tasks through appropriate modifications.
9) UniMatch-DeepLabv3+[38]: Consistent with the approach of UniMatch-PSPNet, the foundational framework is the more robust feature extractor DeepLabv3+. Adapted through appropriate modifications for change detection tasks, this framework ensures better adaptability.

*C. Quantitative analysis*

**GoogleGZ-CD dataset**: While the GoogleGZ-CD dataset may have a smaller scale compared to some other datasets, its unique challenges become apparent when analyzing building features. Firstly, from a supervised perspective, our proposed C2FNet demonstrates strong feature extraction capabilities. Even with the utilization of only 5% labelled data, it achieves an impressive F1 of 78.14%. This result significantly outperforms other semi-supervised methods. Secondly, taking a semi-supervised stance, at the same proportion of labelled samples, our C2F-SemiCD has the ability to glean insights from a substantial volume of unlabeled data. Thirdly, across labelled sample ratios of 5%, 10%, 20%, and 30%, C2F-SemiCD outperforms the runner-up TCNet by margins of 6.18%, 3.93%, 1.55%, and 3.01% respectively. This demonstrates that our C2F-SemiCD excels across all four labelled sample proportions, with its optimal performance at the lowest labelled ratio. This further substantiates that, with respect to the GoogleGZ-CD dataset, our C2F-SemiCD is adept at learning features from a plethora of unlabeled data, thereby showcasing the essence of semi-supervised learning. Overall, the C2F-SemiCD approach proves to be the most effective on the GoogleGZ-CD dataset.

TABLE IV
QUANTITATIVE ACCURACY RESULTS OF SOTA COMPARISON METHODS WITH DIFFERENT LABELLED RATIOS
ON THE LEVIR-CD DATASET. RED HIGHLIGHTS THE BEST VALUES.

| Method | Labelled Ration | | | | | | | | | | | |
|---|---|---|---|---|---|---|---|---|---|---|---|---|
| | 5% | | | | | | 10% | | | | | |
| | F1 | Pre. | Rec. | OA | KC | IoU | F1 | Pre. | Rec. | OA | KC | IoU |
| Super. only (C2FNet) | 83.12 | 95.27 | 73.73 | 98.47 | 82.34 | 71.12 | 86.98 | 95.26 | 80.03 | 98.78 | 86.35 | 76.97 |
| AdvNet | 81.48 | 89.19 | 74.99 | 98.26 | 80.57 | 68.75 | 85.37 | 90.98 | 80.41 | 98.60 | 84.64 | 74.48 |
| S4GAN | 79.49 | 88.57 | 72.10 | 98.10 | 78.51 | 65.96 | 83.46 | 89.19 | 78.43 | 98.42 | 82.64 | 71.62 |
| SemiCDNet | 82.16 | 89.25 | 76.11 | 98.32 | 81.28 | 69.72 | 84.88 | 91.32 | 79.30 | 98.56 | 84.13 | 73.74 |
| SemiCD | 85.03 | 89.17 | 81.25 | 98.54 | 84.26 | 73.95 | 86.75 | 91.85 | 82.19 | 98.72 | 86.08 | 76.60 |
| RCL | 80.01 | 82.26 | 77.87 | 98.02 | 78.96 | 66.68 | 83.78 | 85.79 | 81.86 | 98.39 | 82.93 | 72.09 |
| TCNet | 84.35 | 91.95 | 77.91 | 98.53 | 83.58 | 72.94 | 89.11 | 91.80 | 86.57 | 98.92 | 88.54 | 80.36 |
| UniMatch-PSPNet | 86.13 | 93.37 | 79.93 | 98.69 | 85.45 | 75.64 | 86.10 | 95.24 | 78.56 | 98.71 | 85.43 | 75.59 |
| UniMatch-DeepLabv3+ | 87.10 | 94.08 | 81.08 | 98.69 | 86.41 | 77.15 | 88.05 | 94.51 | 82.42 | 98.78 | 87.42 | 78.66 |
| **C2F-SemiCD** | **89.97** | **91.45** | **88.53** | **98.99** | **89.44** | **81.76** | **90.80** | **92.44** | **89.22** | **99.08** | **90.31** | **83.15** |
| Method | Labelled Ration | | | | | | | | | | | |
| | 20% | | | | | | 30% | | | | | |
| | F1 | Pre. | Rec. | OA | KC | IoU | F1 | Pre. | Rec. | OA | KC | IoU |
| Super. only (C2FNet) | 87.14 | 96.20 | 79.64 | 98.80 | 86.52 | 77.21 | 89.88 | 94.25 | 85.91 | 99.01 | 89.37 | 81.63 |
| AdvNet | 86.62 | 91.34 | 82.37 | 98.70 | 85.94 | 76.40 | 86.85 | 91.36 | 82.76 | 98.72 | 86.18 | 76.75 |
| S4GAN | 85.00 | 93.03 | 78.24 | 98.59 | 84.26 | 73.91 | 86.53 | 92.29 | 81.45 | 98.71 | 85.86 | 76.26 |
| SemiCDNet | 86.25 | 92.15 | 81.06 | 98.68 | 85.56 | 75.82 | 86.92 | 91.25 | 82.98 | 98.73 | 86.25 | 76.87 |
| SemiCD | 87.53 | 91.30 | 84.07 | 98.78 | 86.89 | 77.83 | 87.55 | 91.77 | 83.69 | 98.79 | 86.91 | 77.85 |
| RCL | 85.87 | 86.60 | 85.14 | 98.57 | 85.11 | 75.23 | 86.77 | 86.66 | 86.88 | 98.65 | 86.06 | 76.64 |
| TCNet | 89.95 | 92.21 | 87.80 | 99.00 | 89.43 | 81.74 | 89.92 | 93.38 | 86.71 | 99.01 | 89.40 | 81.68 |
| UniMatch-PSPNet | 87.69 | 94.19 | 82.04 | 98.83 | 87.08 | 78.08 | 87.13 | 93.99 | 81.20 | 98.78 | 86.49 | 77.19 |
| UniMatch-DeepLabv3+ | 88.46 | 94.62 | 83.05 | 98.82 | 87.84 | 79.31 | 88.63 | 94.73 | 83.27 | 98.99 | 88.11 | 79.59 |
| **C2F-SemiCD** | **91.16** | **93.26** | **89.15** | **99.12** | **90.69** | **83.75** | **91.58** | **93.78** | **89.47** | **99.16** | **91.14** | **84.46** |
| Super. only | (C2FNet): 100% → | | | | | | **91.83** | **93.69** | **90.04** | **99.18** | **91.40** | **84.89** |

**WHU-CD dataset**: First, when the proportion of labelled samples is 5% and 10%, C2F-SemiCD achieves the third score, which is 2.72% and 2% lower than the first-place UniMatch-DeepLabv3+, but still outperforms TCNet, which achieves the second place on GoogleGZ-CD dataset. There are two reasons for our analysis: the WHU-CD dataset has obvious change characteristics between the images of the two time phases, and the background is relatively simple, so it is relatively easy to learn the building features in it. PSPNet and DeepLabv3+ methods widely used in the field of computer vision have good feature extraction ability. Second, when the proportion of labelled samples is 20% and 30%, our proposed C2F-SemiCD method achieves the first performance, surpassing the second-place UniMatch-DeepLabv3+ method by 0.81% and 2.38%. When the proportion of labelled samples is 5%, 10%, 20%, and 30%, our semi-supervised C2F-SemiCD method can still improve 7.75%, 6.39%, 4.6%, 4.85% compared with the supervised method C2FNet. This also further proves that for the WHU-CD dataset, our proposed C2F-SemiCD can learn features from a large number of unlabeled data, which can achieve the original intention of semi-supervised learning designed by us. In general, the C2F-SemiCD method can achieve relatively good results on the WHU-CD dataset

**LEVIR-CD dataset:** The LEVIR-CD dataset is relatively large, and the overall detection accuracy value is significantly higher than the GoogleGZ-CD and WHU-CD datasets. In the case of 5%, 10%, 20% and 30% labelled samples, our proposed C2F-SemiCD method outperforms the second place UniMatch-DeepLabv3+ or TCNet method by 2.87%, 1.69%, 1.21% and 1.66%, respectively, and reaches the first place. However, this difference is the largest when the labelled samples are 5%, which directly proves that our C2F-SemiCD method can learn better unlabeled data features with a small number of labelled samples. In the case that the accuracy of the fully supervised C2F-SemiCD method is lower than that of the partially semi-supervised method, our proposed C2F-SemiCD can learn the features of buildings well in a large amount of unlabeled data. When the proportion of labelled samples is 5%, 10%, 20%, and 30%, the proposed C2F-SEMICD can learn the features of buildings well. Compared with the supervised method C2FNet, the C2F-SemiCD method can still improve by 6.85%, 3.82%, 4.02%, and 1.7%, thus achieving the first place, which clearly proves that C2F-SemiCD has a better semi-supervised learning effect. The metrics we focus on are F1, IoU, and KC, where higher numbers indicate better detection performance of the model. In general, our proposed C2F-SemiCD can achieve the best results on the LEVIR-CD dataset.

*D. Qualitative Visualization*

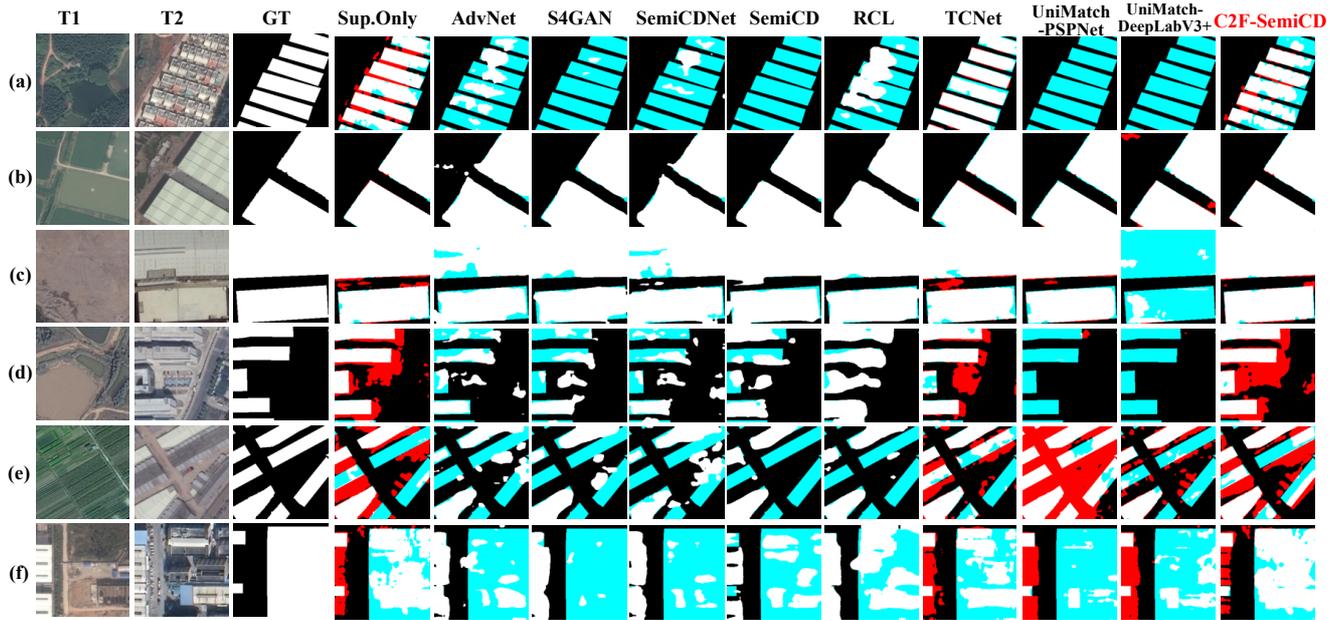

**Fig. 6.** Qualitative evaluation results of SOTA comparison methods with a 5% labelled ratio on the GoogleGZ-CD dataset. TP (white), TN (black), FP (red), and FN (blue).

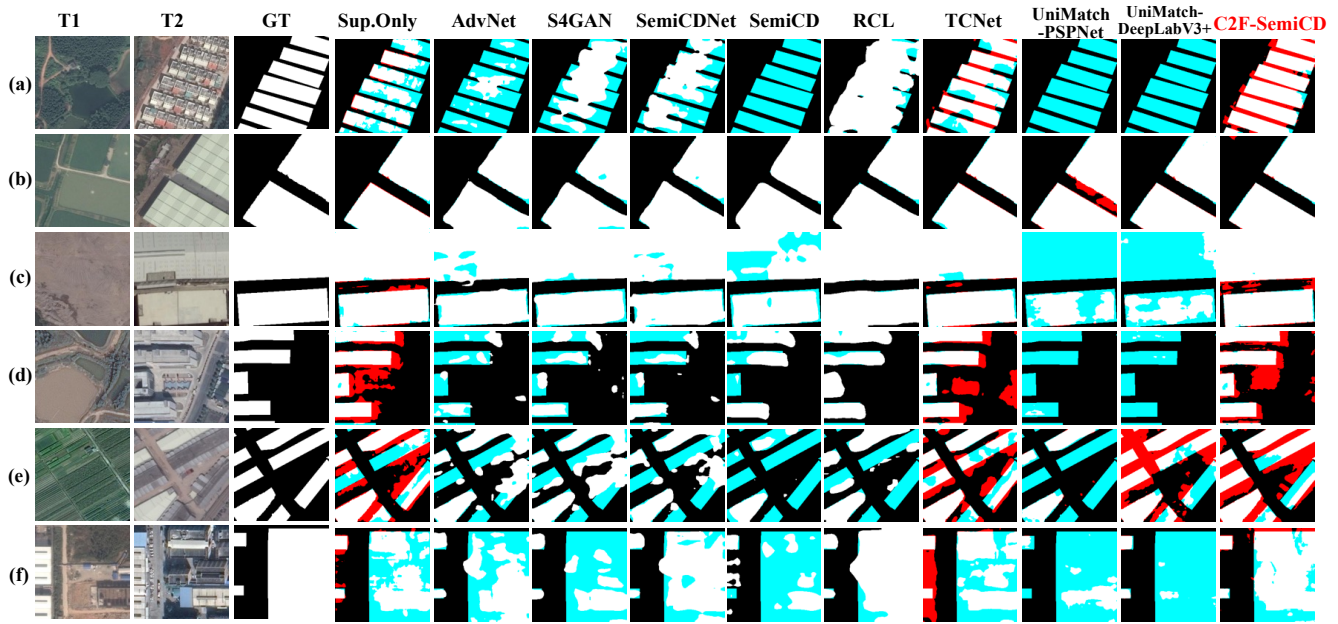

**Fig. 7.** Qualitative evaluation results of SOTA comparison methods with a 10% labelled ratio on the GoogleGZ-CD dataset.

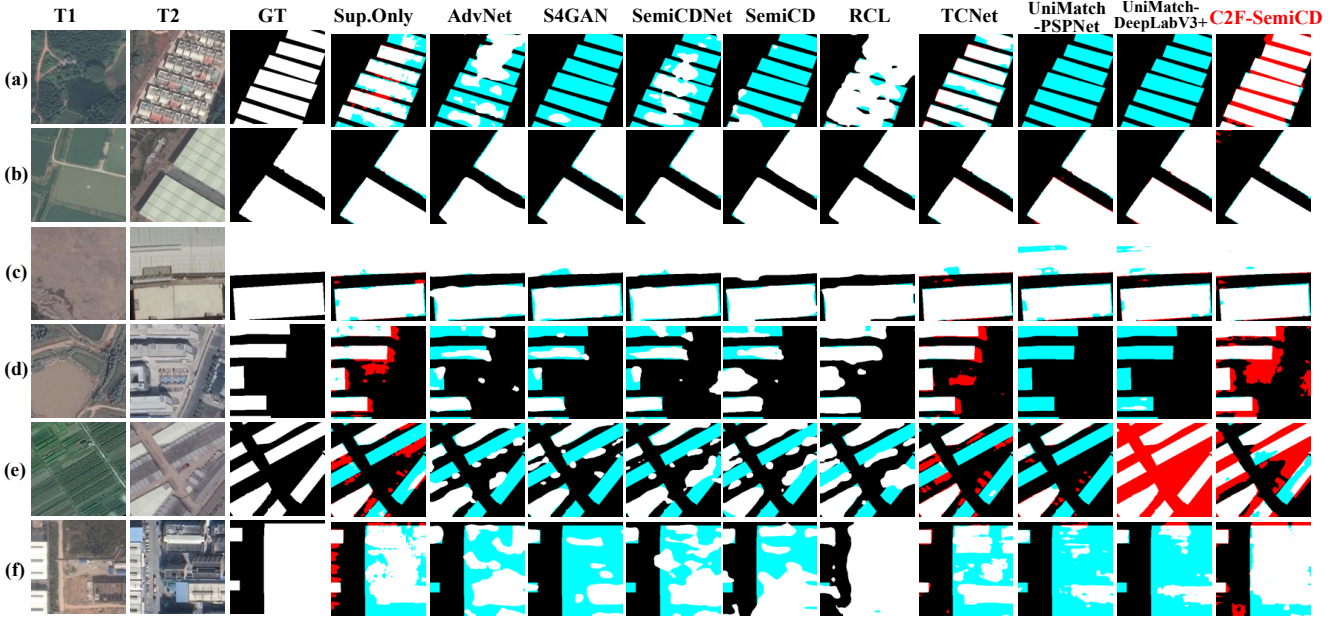

**Fig. 8.** Qualitative evaluation results of SOTA comparison methods with a 20% labelled ratio on the GoogleGZ-CD dataset.

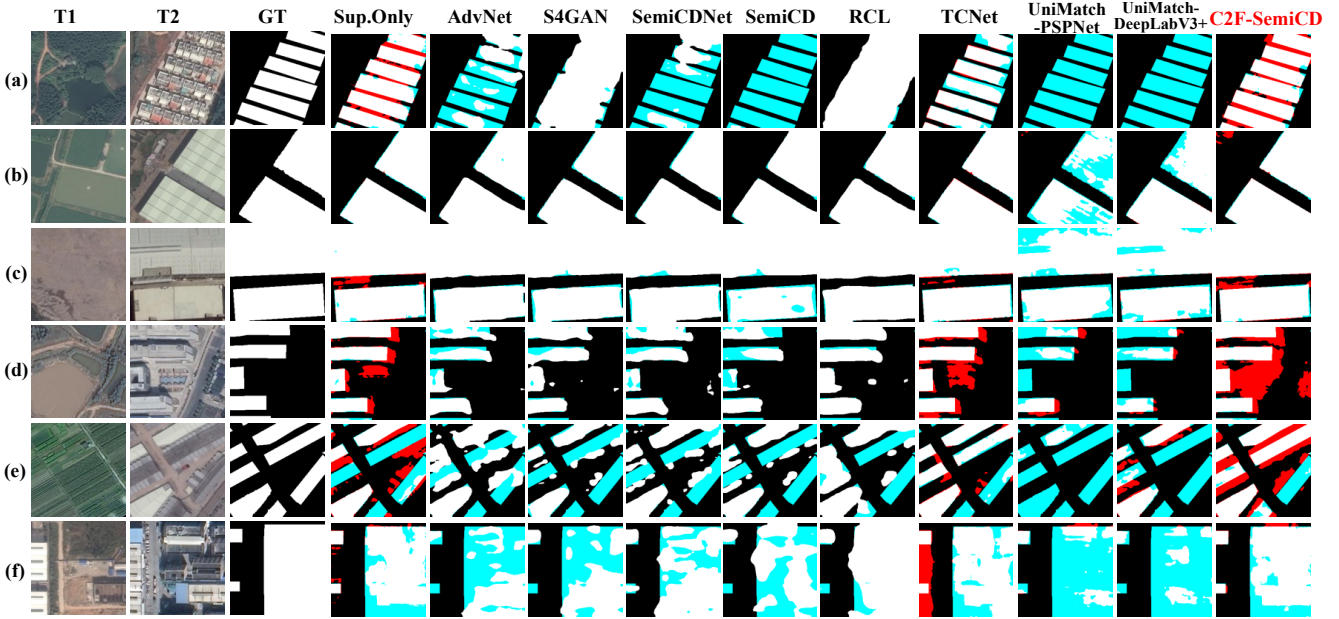

**Fig. 9.** Qualitative evaluation results of SOTA comparison methods with a 30% labelled ratio on the GoogleGZ-CD dataset

**GoogleGZ-CD dataset**: Fig.6-Fig. 9 shows the qualitative visualization results of four annotation proportions (5%, 10%, 20%, 30%) of our proposed C2F-SemiCD compared with all change detection methods of SOTA on the GoogleGZ-CD dataset. To make the visualization more intuitive, we use white for TP, black for TN, red for FP, and light blue for FN. By observing the visualization, we can draw the following conclusions: first, with the increase of the proportion of labelled samples, the visualization effect is improved to a certain extent, and the blue and red parts are decreased to a certain extent. Secondly, the effect of the proposed supervised method C2FNet is also more significant than that of other semi-supervised methods, which is consistent with the quantitative results. Third, the results of the proposed semi-supervised method C2F-SemiCD remain the best overall, with relatively few blue and red parts (i.e., relatively few missed and misdetected parts) on the visualized graphs. It is worth noting that we selected 6 representative images from 300 test images, and the calculated accuracy is also the overall average accuracy of 300 test images, so the visualization results cannot be strictly compared with the quantization results. In general, the proposed semi-supervised method C2F-SemiCD has the best visualization effect on the GoogleGZ-CD dataset.

**WHU-CD dataset**: Fig. 10-Fig. 13 shows the qualitative visualization results of our proposed C2F-SemiCD compared with all change detection methods of SOTA on the WHU-CD dataset with four annotation proportions (5%, 10%, 20%, 30%). To make the visualization more intuitive, we use white for TP, black for TN, red for FP, and light blue for FN. By observing the visualization graph, we can draw the following conclusions: First, with the increase of the proportion of labelled samples, the visualization effect is significantly improved compared with GoogleGZ-CD, especially the missed detection of the blue part is reduced by a lot. Second, the semi-supervised method C2F-SemiCD and the supervised method C2FNet are consistent with the quantitative results, and the red false detection part and the blue missed detection part are visually degraded. Thirdly, compared with other semi-supervised methods, the proposed semi-supervised method C2F-SemiCD performs relatively better on the blue missed detection part, while the red false detection part is slightly more obvious on individual images. In general, the proposed semi-supervised method C2F-SemiCD performs better in visualization on the WHU-CD dataset.

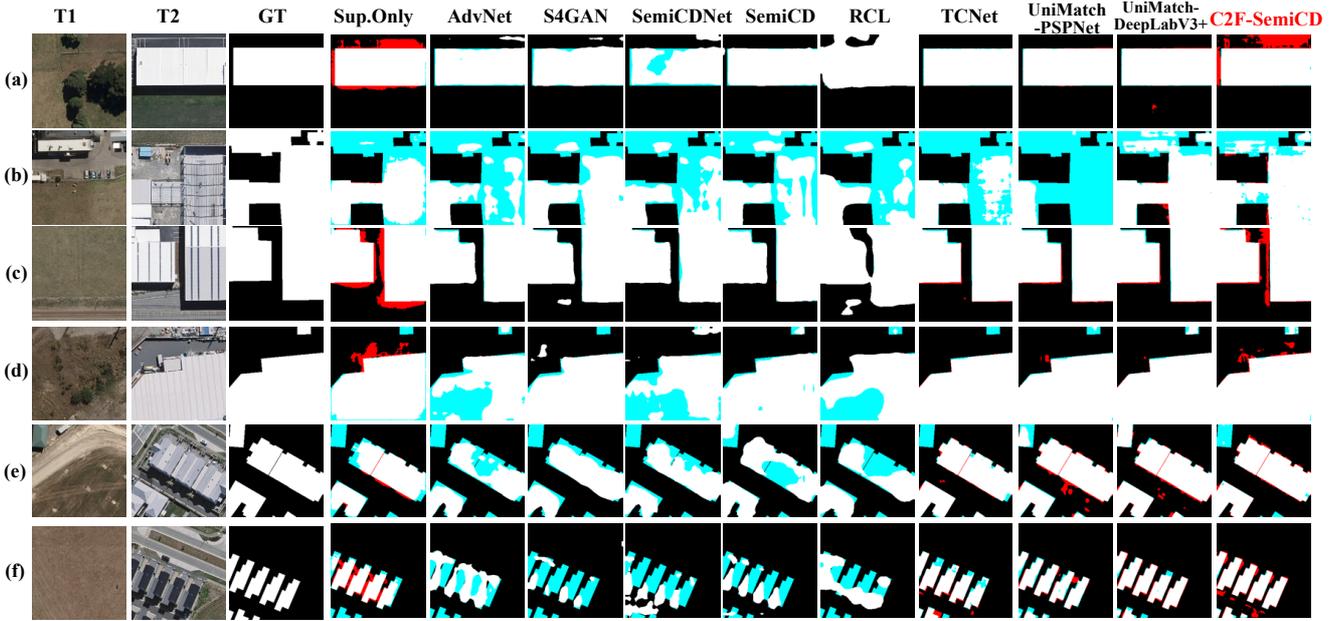

**Fig. 10.** Qualitative evaluation results of SOTA comparison methods with a 5% labelled ratio on the WHU-CD dataset.

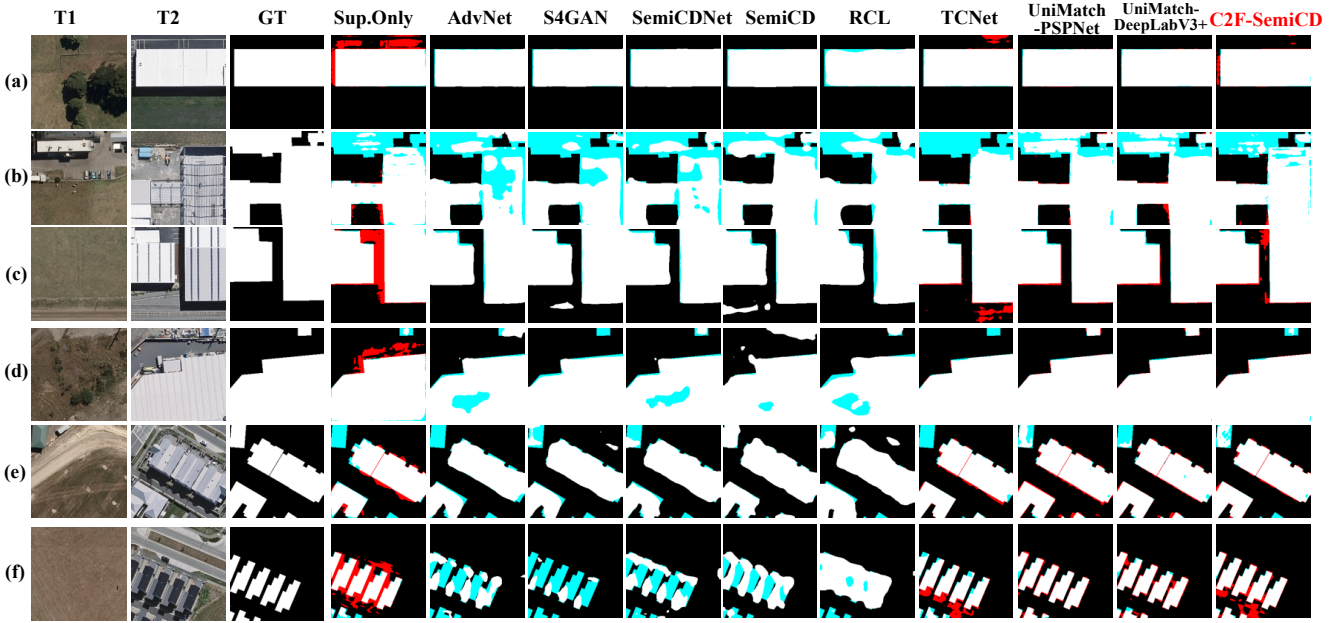

**Fig. 11.** Qualitative evaluation results of SOTA comparison methods with a 10% labelled ratio on the WHU-CD dataset

**LEVIR-CD dataset**: Fig. 14-Fig. 17 shows the qualitative visualization results of four labelled ratios (5%, 10%, 20%, 30%) of our proposed C2F-SemiCD compared with all change detection methods of SOTA on the LEVI-CD dataset. To make the visualization more intuitive, we use white for TP, black for TN, red for FP, and light blue for FN. By observing the visualization, we can draw the following conclusions: First, all of these methods perform well in the detection of small buildings, and the missed detection phenomenon is relatively obvious in large buildings;

Secondly, the semi-supervised method C2F-SemiCD proposed in this paper has a relatively obvious improvement in the missed detection of large buildings compared with other semi-supervised methods. Of course, there are some very obvious missed detection areas which are also common difficulties in all methods. Thirdly, the proposed semi-supervised method C2F-SemiCD significantly improves the visualization effect compared with other supervised methods C2FNet. In general, the proposed semi-supervised method C2F-SemiCD has the best visualization performance on the LEVIR-CD dataset. Because both WHU-CD and LEVIR-CD datasets are datasets for building change detection, we want to explore the robustness of the semi-supervised method in the form of a cross-dataset. First, the verification is to learn the features of LEVIR-CD while training the WHU-CD dataset. For ease of understanding, we denote training on the labelled dataset WHU-CD and the unlabeled set LEVIR-CD by {WHU (supervised), LEVIR (un-supervised)}→WHU and testing on the WHU-CD dataset. By observing TABLE V, we can draw the following conclusions: First, when the proportion of labelled samples is 5%, 10%, 20%, compared with the F1 score of 85.63%, 86.58%, 90.07% on the semi-supervised training (experiment 2) on the unlabeled dataset WHU-CD, the F1 score on the trans-LEVI-CD dataset is 88.09%, 89.74%, 90.85%, and the F1 score on the trans-LEVI-CD dataset is 88.09%, 89.74%, 90.85%. The effect is more significant, which directly proves that semi-supervised learning without labelled data on building cross-dataset is effective. Second, when the WHU-CD dataset has only a small number of labelled samples, it can learn the building features on the unlabeled dataset LEVIR-CD and has a significant improvement effect. When the proportion of labelled samples is 5%, 10%, 20%, and 30%, through cross-dataset semi-supervised learning, Compared with the supervised method C2FNet, our semi-supervised C2F-SemiCD method can improve the F1 value by 10.21%, 9.55%, 5.38%, 3.6%. Third, in this cross-dataset case, the proposed C2F-SemiCD method outperforms all other leading-edge eight semi-supervised change detection methods. In general, the proposed semi-supervised C2F-SemiCD method achieves significant results on the cross-dataset {WHU (supervised), LEVIR (un-supervised)}→WHU experiments.

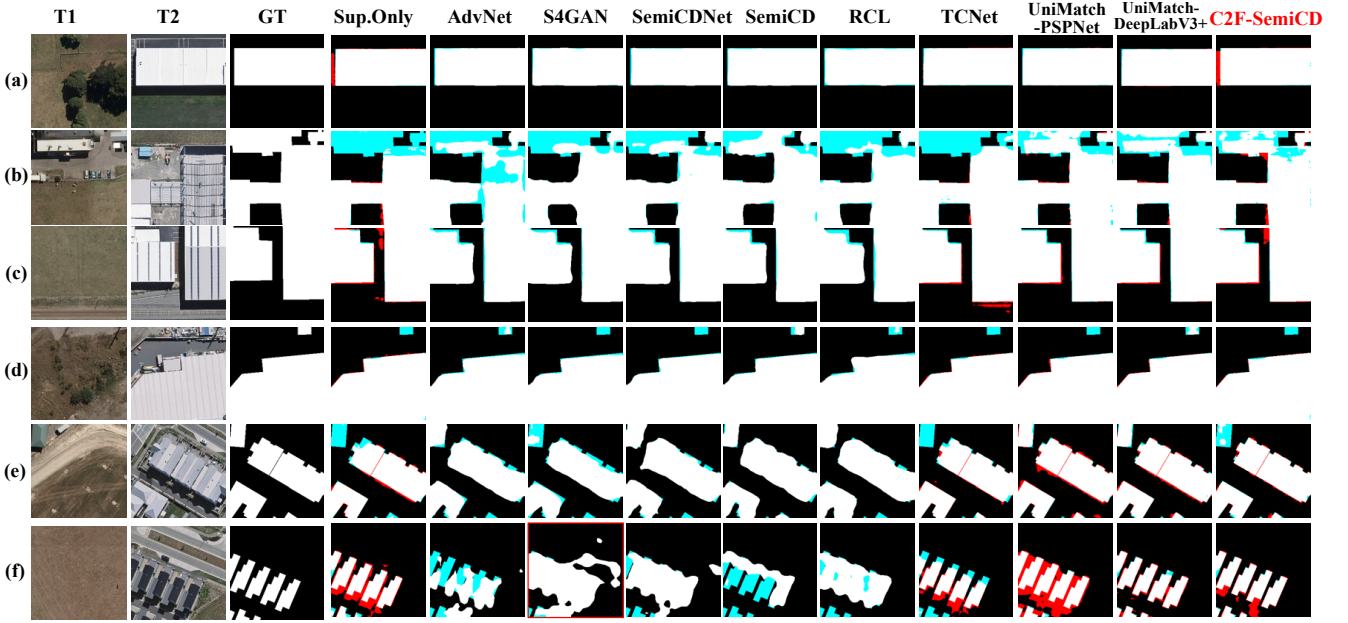

**Fig. 12.** Qualitative evaluation results of SOTA comparison methods with a 20% labelled ratio on the WHU-CD dataset.

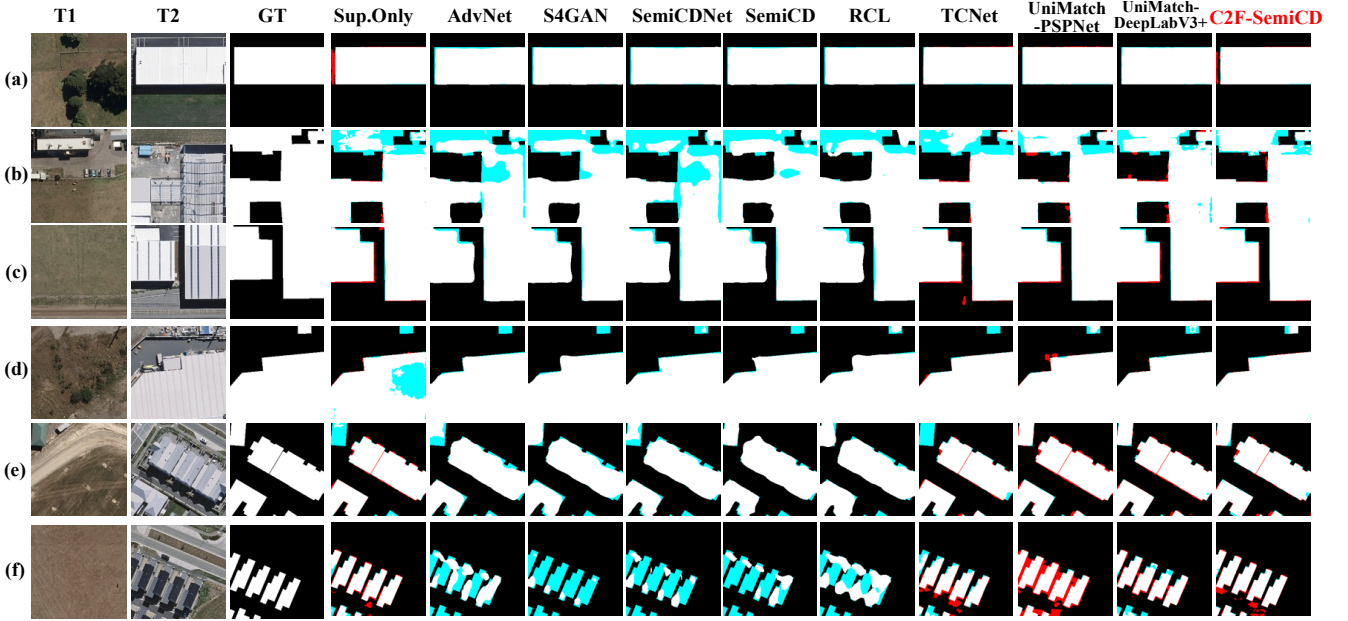

**Fig. 13.** Qualitative evaluation results of SOTA comparison methods with a 30% labelled ratio on the WHU-CD dataset.

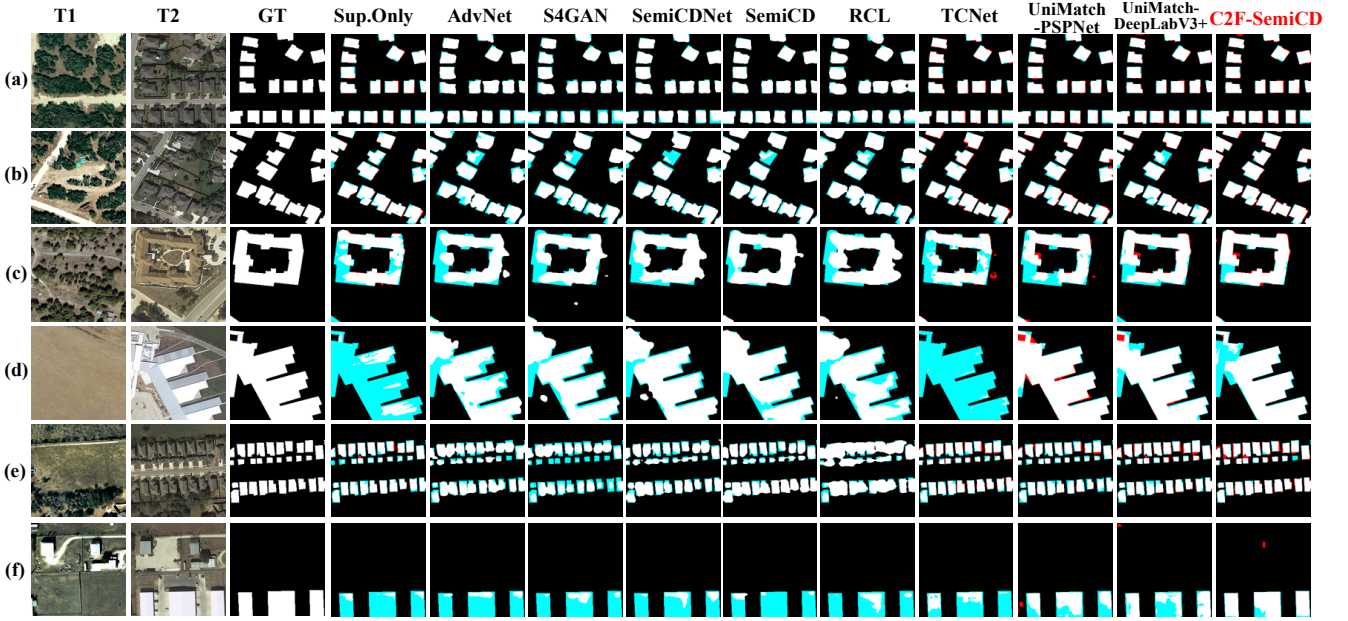

**Fig. 14.** Qualitative evaluation results of SOTA comparison methods with a 5% labelled ratio on the LEVIR-CD dataset.

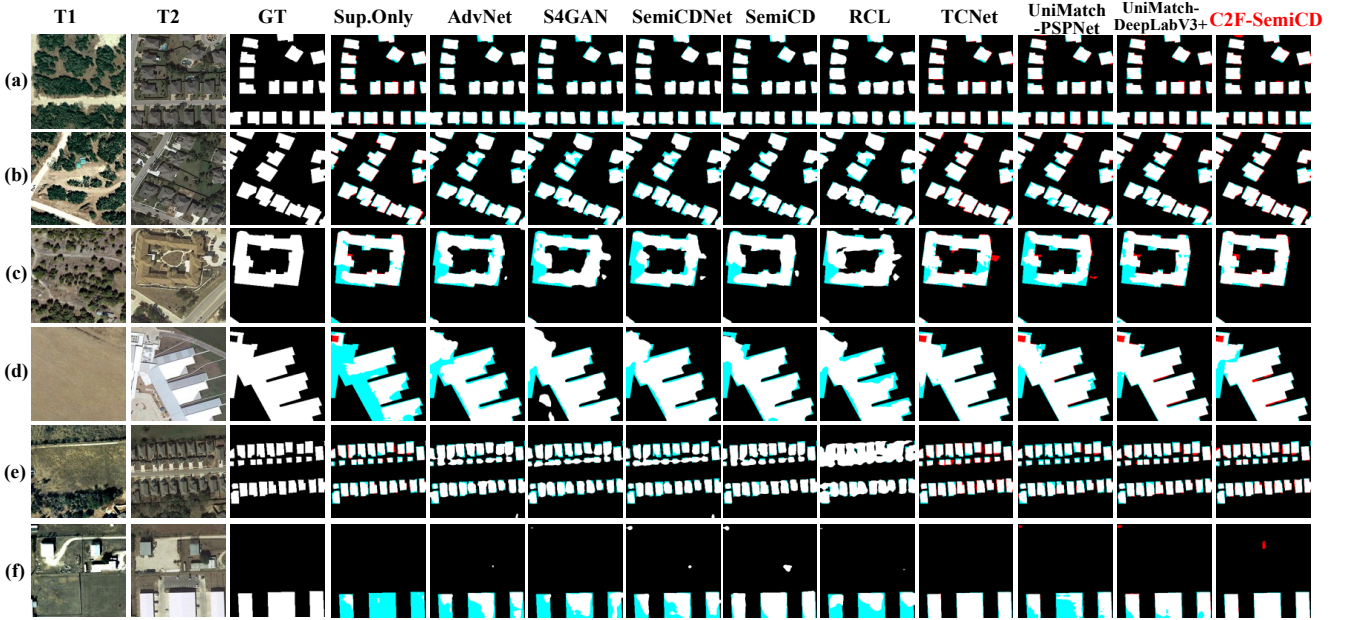

**Fig. 15.** Qualitative evaluation results of SOTA comparison methods with a 10% labelled ratio on the LEVIR-CD dataset.

Secondly, the verification is to learn the characteristics of WHU-CD when training the LEVIR-CD dataset. For ease of understanding, we denote by {LEVIR(supervised), WHU(un-supervised)}→LEVIR training on the labelled dataset LEVIR-CD and the unlabeled dataset WHU-CD and testing on the LEVIR-CD dataset. By observing TABLE VI, we can draw the following conclusions: First, when the proportion of labelled samples is 5%, 10%, 20%, and 30%, the F1 score on the cross-WHU-CD dataset is 87.97%, compared with 89.97%, 90.80%, 91.16%, 91.58% on the semi-supervised training (Experiment 3) on the unlabeled dataset LEVI-CD. 89.74%, 90.85%, 91.36%. Although the latter is relatively low, we analyze the possible reason that the WHU-CD dataset is less than the LEVIR-CD dataset, so there are fewer features when learning unlabeled datasets. Second, when the LEVIR-CD dataset has only a small number of labelled samples, it can learn the building features on the WHU-CD dataset without labels and has a significant improvement effect. When the proportion of labelled samples is 5%, 10%, 20%, and 30%, through semi-supervised learning across datasets, it can achieve a significant improvement. Compared with the supervised method C2FNet, our semi-supervised C2F-SemiCD method can improve the F1 score by 4.85%, 2.66%, 3.64%, 1.48%. Third, in this cross-dataset case, the proposed C2F-

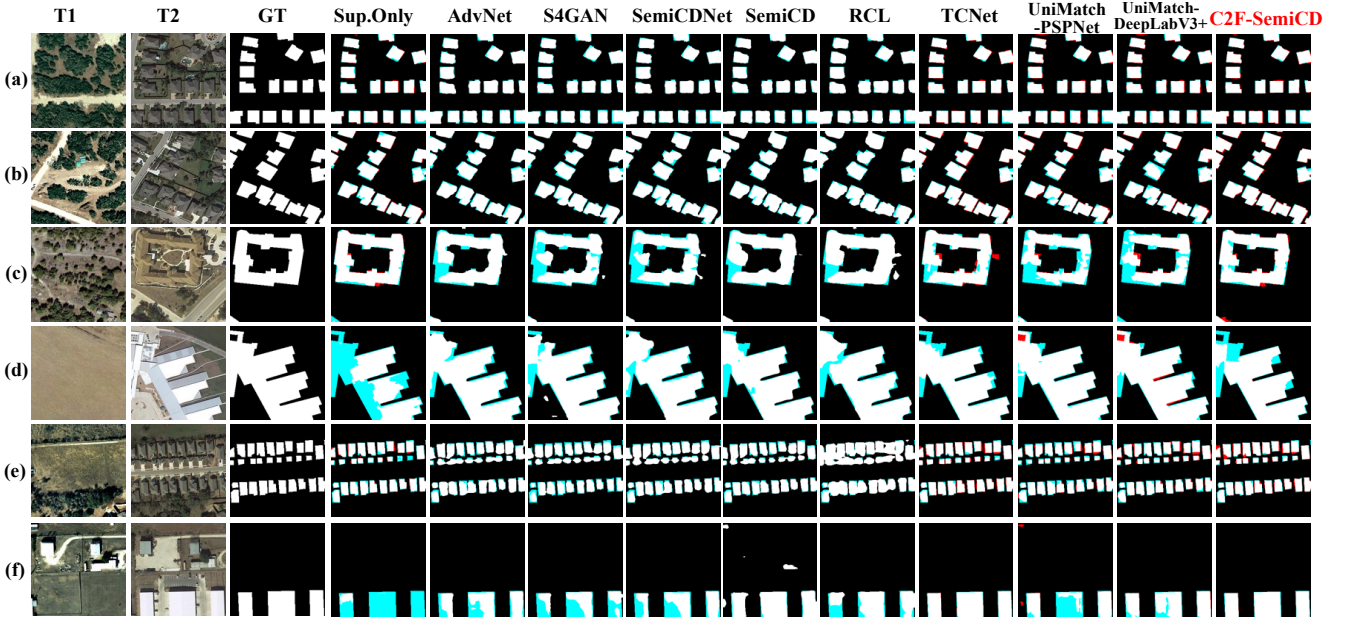

**Fig. 16.** Qualitative evaluation results of SOTA comparison methods with a 20% labelled ratio on the LEVIR-CD dataset.

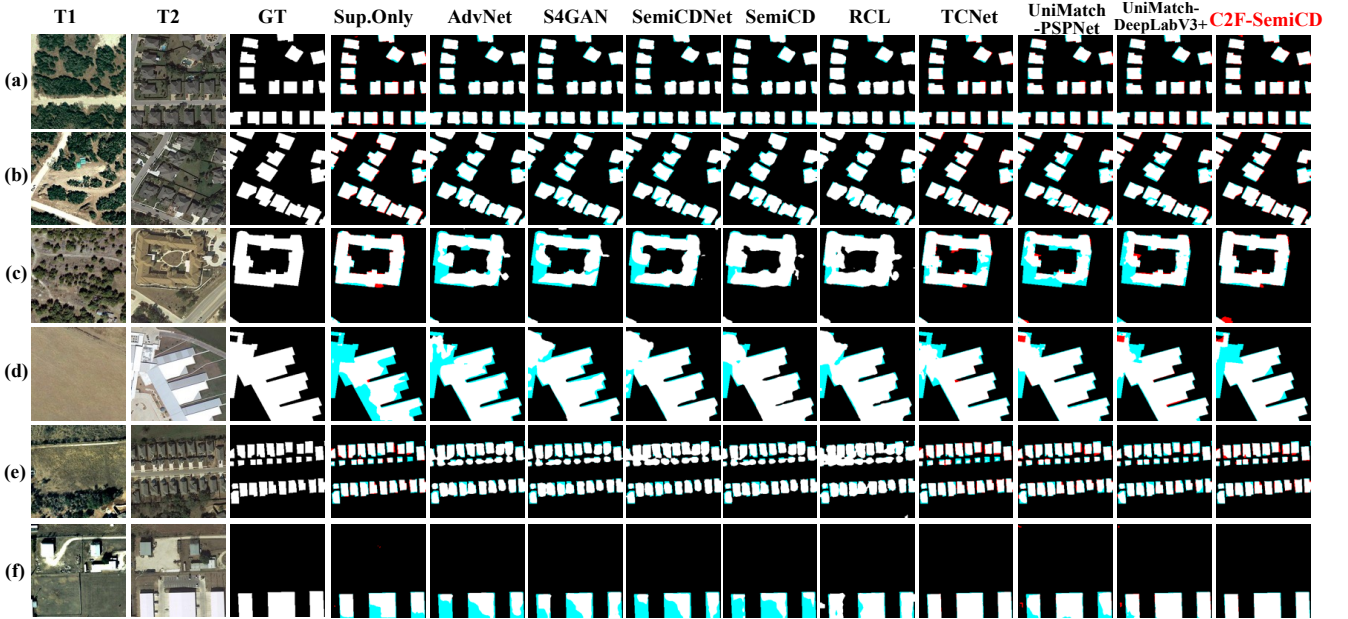

**Fig. 17.** Qualitative evaluation results of SOTA comparison methods with a 30% labelled ratio on the LEVIR-CD dataset.

SemiCD method outperforms all other leading-edge eight semi-supervised change detection methods. In general, the proposed semi-supervised C2F-SemiCD method achieves significant results on the cross-dataset {LEVIR(supervised), WHU(un-supervised)}→LEVIR experiments.

*E. Ablation Study*

*Ablation on modules in C2FNet*: In order to verify the effect of the global contextual module (GCM), Refine Module, Aggregation_Init and Aggregation_Final of the proposed C2FNet in C2F-SemiCD, we verified the effect of each module on 5% and 10% label ratio of GoogleGZ-CD data. As shown in TABLE VII, we can see that the larger the difference between the number in each row and the number in the last row, the better the performance of the module. Aggregation_Final works best when the number of tags is 5%. The global context module GCM_C'D'E' works best when the proportion of labels is 10%. There are two stages with GCM in the network. By comparing GCM_ABC, GCM_ C'D'E', and all GCM, it can be seen that the gap between C2F-SemiCD and GCM_ABC is larger when GCM_ABC and GCM_C'D'E' are removed. It means that the network works better when it has both GCM_ABC and GCM_C' D'E'. In general, each of the proposed modules can play a key role in the process of feature extraction.

*Ablation on initial supervised training times*: To validate the hyperparameter of the initial supervised training times in semi-supervised learning, we verified the dissimilarity between the number of supervised times of 5, 10, 15, and 20 on the 5% and 10% label proportions of the WHU-CD dataset, as shown in Fig.18. Through a large number of experiments, we find that: First, it is not that the higher the number of supervised training, the higher the accuracy; Second, when the number of supervised training is 15, both accuracies are the highest; Third, our previous experiments used a coefficient of 5 instead of 15. The main reason is that a small amount of supervised training can highlight the original intention of semi-supervised learning (learning features from a large amount of unlabeled data), rather than relying too much on supervised learning.

*Ablation on semi-supervised coefficients in the loss function*: In order to verify the hyperparameters of the semi-supervised coefficients in the loss function, we verified the difference between coefficients of 0.1 to 0.9 on the 5% and 10% label proportion of the WHU-CD dataset, as shown in Fig.19. The coefficient used in our previous experiments was 0.2. Through a large number of experiments, we find that: First, the influence of this coefficient on the accuracy of the model is not particularly large; Second, the performance of different label proportions is not the same, and it is difficult to find the optimal solution; Third, perhaps we can try to mitigate some of the volatility caused by this coefficient by repeating the experiment many times and taking the average.

*Ablation on training and inference time*: To verify the operational efficiency of our method, we list the training and inference time of all models on GoogleGZ-CD in Fig.20. Compared with all other semi-supervised methods, our proposed C2F-SemiCD ranks third in training time and first in inference time. When we take the accuracy of our model into account, we can see that our accuracy is much higher than AdvNet and SemiCDNet. Therefore, our proposed C2F-SemiCD has very significant advantages in training and inference.

IV. CONCLUSION

In this paper, aiming at the problem that many supervised change detection methods of deep learning can only learn and recognize the patterns of change features from a large number of labelled images, we propose a coarse-grained to fine-grained semi-supervised change detection method based on consistency regularization(C2F-SemiCD) and a coarse-to-fine change detection network with a multi-scale attention mechanism (C2FNet). The extraction of change features from coarse-grained to fine-grained is completed "step by step". We conducted extensive experiments and detailed ablation studies on four ratios (5%, 10%, 20%, 30%) of the three datasets, including experiments across datasets. Through extensive visualizations, the effectiveness of our method is demonstrated intuitively from

TABLE V
CROSS-VALIDATION EXPERIMENTS ON {WHU (SUPERVISED), LEVIR (UN-SUPERVISED)}→WHU. RED HIGHLIGHTS THE BEST VALUES.

| Method | Labelled Ration | | | | | | | | | | | |
|---|---|---|---|---|---|---|---|---|---|---|---|---|
| | 5% | | | | | | 10% | | | | | |
| | F1 | Pre. | Rec. | OA | KC | IoU | F1 | Pre. | Rec. | OA | KC | IoU |
| Super. only (C2FNet) | 77.88 | 85.80 | 71.29 | 98.39 | 77.05 | 63.77 | 80.19 | 81.72 | 78.72 | 98.46 | 79.39 | 66.93 |
| AdvNet | 71.45 | 75.61 | 67.72 | 97.85 | 70.34 | 55.58 | 77.96 | 78.35 | 77.57 | 98.26 | 77.05 | 63.88 |
| S4GAN | 46.91 | 82.36 | 32.79 | 97.05 | 45.68 | 30.64 | 57.53 | 68.69 | 49.49 | 97.10 | 56.07 | 40.38 |
| SemiCDNet | 72.87 | 78.51 | 68.00 | 97.99 | 71.84 | 57.33 | 76.72 | 75.71 | 77.76 | 98.13 | 75.75 | 62.24 |
| SemiCD | 74.44 | 76.73 | 72.28 | 98.03 | 73.42 | 59.29 | 80.27 | 81.22 | 79.33 | 98.45 | 79.46 | 67.04 |
| RCL | 70.88 | 71.56 | 70.21 | 97.71 | 69.69 | 54.89 | 75.26 | 70.41 | 80.83 | 97.89 | 74.17 | 60.33 |
| TCNet | 85.97 | 91.78 | 80.86 | 98.74 | 85.31 | 75.40 | 86.65 | 89.91 | 85.51 | 98.03 | 87.14 | 79.24 |
| UniMatch -PSPNet | 42.37 | 90.85 | 27.63 | 97.02 | 41.28 | 26.88 | 76.65 | 82.68 | 71.44 | 98.27 | 75.76 | 62.14 |
| UniMatch -DeepLabv3+ | 50.35 | **94.59** | 34.30 | 97.32 | 49.28 | 33.64 | 76.27 | 83.81 | 69.97 | 98.27 | 75.38 | 61.64 |
| **C2F-SemiCD** | **88.09** | 90.55 | **85.77** | **98.89** | **87.51** | **78.72** | **89.74** | **92.19** | **87.42** | **99.04** | **89.24** | **81.39** |
| Method | Labelled Ration | | | | | | | | | | | |
| | 20% | | | | | | 30% | | | | | |
| | F1 | Pre. | Rec. | OA | KC | IoU | F1 | Pre. | Rec. | OA | KC | IoU |
| Super. only (C2FNet) | 85.47 | 89.35 | 81.92 | 98.90 | 84.90 | 74.63 | 88.00 | **93.57** | 83.06 | 99.10 | 87.54 | 78.58 |
| AdvNet | 82.38 | 79.33 | 85.68 | 98.55 | 81.63 | 70.04 | 84.29 | 83.00 | 85.62 | 98.73 | 83.63 | 72.84 |
| S4GAN | 74.18 | 84.80 | 65.93 | 98.18 | 73.25 | 58.96 | 76.80 | 85.12 | 69.97 | 98.32 | 75.94 | 62.34 |
| SemiCDNet | 81.26 | 75.34 | 88.20 | 98.39 | 80.42 | 68.44 | 85.41 | 82.60 | 88.42 | 98.80 | 84.79 | 74.54 |
| SemiCD | 83.31 | 78.69 | 88.51 | 98.59 | 82.58 | 71.39 | 84.51 | 80.64 | 88.77 | 98.71 | 83.84 | 73.17 |
| RCL | 83.35 | 80.44 | 86.47 | 98.63 | 82.63 | 71.45 | 86.83 | 86.57 | 87.09 | 98.95 | 86.28 | 76.72 |
| TCNet | 89.16 | 91.85 | 86.62 | 98.99 | 88.63 | 80.44 | 90.25 | 92.56 | 88.06 | 99.09 | 89.77 | 82.24 |
| UniMatch -PSPNet | 81.29 | 77.84 | 85.05 | 98.45 | 80.48 | 68.47 | 87.97 | 87.68 | 88.26 | 99.04 | 87.47 | 78.52 |
| UniMatch -DeepLabv3+ | 80.06 | **93.69** | 69.90 | 98.62 | 79.36 | 66.75 | 88.66 | 92.01 | 85.55 | 99.13 | 88.21 | 79.63 |
| **C2F-SemiCD** | **90.85** | **92.92** | **88.86** | **99.14** | **90.40** | **83.23** | **91.60** | 93.52 | **89.75** | **99.21** | **91.18** | **84.50** |

TABLE VI
CROSS-VALIDATION EXPERIMENTS ON { LEVIR (SUPERVISED), WHU(UN-SUPERVISED)}→LEVIR. RED HIGHLIGHTS THE BEST VALUES.

| Method | Labelled Ration | | | | | | | | | | | |
|---|---|---|---|---|---|---|---|---|---|---|---|---|
| | 5% | | | | | | 10% | | | | | |
| | F1 | Pre. | Rec. | OA | KC | IoU | F1 | Pre. | Rec. | OA | KC | IoU |
| Super. only (C2FNet) | 83.12 | 95.27 | 73.73 | 98.47 | 82.34 | 71.12 | 86.98 | **95.26** | 80.03 | 98.78 | 86.35 | 76.97 |
| AdvNet | 80.23 | 83.79 | 76.96 | 98.07 | 79.22 | 66.99 | 83.33 | 89.49 | 77.96 | 98.41 | 82.50 | 71.42 |
| S4GAN | 79.49 | 87.61 | 72.74 | 98.09 | 78.49 | 65.96 | 83.33 | 90.31 | 77.36 | 98.42 | 82.51 | 71.43 |
| SemiCDNet | 79.78 | 90.21 | 71.51 | 98.15 | 78.82 | 66.36 | 83.16 | 91.11 | 76.49 | 98.42 | 82.34 | 71.17 |
| SemiCD | 85.00 | 88.44 | 81.82 | 98.53 | 84.23 | 73.91 | 86.45 | 89.06 | 93.99 | 98.66 | 85.75 | 76.14 |
| RCL | 80.26 | 82.22 | 78.39 | 98.04 | 79.23 | 67.03 | 84.19 | 84.88 | 83.51 | 98.40 | 83.35 | 72.69 |
| TCNet | 87.01 | 91.05 | 83.31 | 98.81 | 86.38 | 77.00 | 87.61 | 90.40 | 83.11 | 98.77 | 85.96 | 76.37 |
| UniMatch -PSPNet | 85.47 | 91.73 | 80.02 | 98.61 | 84.75 | 74.63 | 87.40 | 91.66 | 83.51 | 98.77 | 86.75 | 77.61 |
| UniMatch -DeepLabv3+ | 80.63 | **95.67** | 69.67 | 98.29 | 79.76 | 67.54 | 86.06 | 93.20 | 79.93 | 98.68 | 85.37 | 75.53 |
| **C2F-SemiCD** | **87.97** | **90.23** | **85.82** | **98.87** | **87.38** | **78.53** | **89.64** | **91.71** | **87.66** | **99.03** | **89.13** | **81.23** |
| Method | Labelled Ration | | | | | | | | | | | |
| | 20% | | | | | | 30% | | | | | |
| | F1 | Pre. | Rec. | OA | KC | IoU | F1 | Pre. | Rec. | OA | KC | IoU |
| Super. only (C2FNet) | 87.14 | **96.20** | 79.64 | 98.80 | 86.52 | 77.21 | 89.88 | **94.25** | 85.91 | 99.01 | 89.37 | 81.63 |
| AdvNet | 84.99 | 89.80 | 80.68 | 98.55 | 84.23 | 73.90 | 86.05 | 90.13 | 82.33 | 98.64 | 85.34 | 75.52 |
| S4GAN | 84.37 | 91.07 | 78.60 | 98.52 | 83.60 | 72.97 | 84.28 | 91.11 | 78.41 | 98.51 | 83.51 | 72.84 |
| SemiCDNet | 84.86 | 91.96 | 78.78 | 98.57 | 84.11 | 73.70 | 85.34 | 90.86 | 80.44 | 98.59 | 84.60 | 74.42 |
| SemiCD | 86.27 | 90.44 | 82.47 | 98.66 | 85.57 | 75.86 | 85.55 | 91.89 | 80.61 | 98.65 | 85.17 | 75.25 |
| RCL | 85.81 | 86.64 | 85.00 | 98.57 | 85.06 | 75.15 | 86.53 | 87.85 | 85.25 | 98.65 | 85.82 | 76.25 |
| TCNet | 89.04 | 92.12 | 86.16 | 98.98 | 88.51 | 80.25 | 89.39 | 92.13 | 86.82 | 99.01 | 88.88 | 80.82 |
| UniMatch -PSPNet | 88.33 | 93.41 | 83.78 | 98.87 | 87.74 | 79.10 | 88.55 | 92.98 | 84.52 | 98.89 | 87.97 | 79.45 |
| UniMatch -DeepLabv3+ | 87.79 | 94.04 | 82.32 | 98.83 | 87.18 | 78.24 | 87.89 | 91.94 | 84.18 | 98.82 | 87.27 | 78.39 |
| **C2F-SemiCD** | **90.78** | **92.66** | **88.98** | **99.13** | **90.33** | **83.12** | **91.36** | **93.48** | **89.33** | **99.19** | **90.93** | **84.09** |

TABLE VII
ABLATION STUDY OF SOME MODULES IN C2FNET ON GOOGLEGZ-CD WITH 5% AND 10% LABELLED DATA.

| Model | 5% | | | | | | 10% | | | | | |
|---|---|---|---|---|---|---|---|---|---|---|---|---|
| | F1 | Pre. | Rec. | OA | KC | IoU | F1 | Pre. | Rec. | OA | KC | IoU |
| w/o GCM_ABC | 80.71 | 78.81 | 82.71 | 90.28 | 74.22 | 67.66 | 81.92 | 79.34 | 84.67 | 90.81 | 75.77 | 69.37 |
| w/o GCM_C'D'E' | 79.81 | 76.94 | 82.90 | 89.69 | 72.90 | 66.40 | 81.32 | 77.75 | 85.23 | 90.38 | 74.85 | 68.52 |
| w/o all GCM | 78.33 | 76.81 | 79.92 | 89.13 | 71.09 | 64.38 | 81.49 | 78.48 | 84.73 | 90.54 | 75.15 | 68.76 |
| w/o Refine | 79.94 | 76.53 | 83.67 | 89.68 | 73.01 | 66.58 | 81.62 | 79.55 | 83.72 | 90.34 | 74.19 | 68.53 |
| w/o Agg_Init | 80.80 | 81.02 | 80.76 | 90.11 | 74.00 | 67.33 | 81.61 | 78.45 | 83.81 | 90.33 | 75.84 | 69.37 |
| w/o Agg_Fin | 78.42 | 76.93 | 79.96 | 89.18 | 71.20 | 64.50 | 81.73 | 78.33 | 85.44 | 90.61 | 75.43 | 69.10 |
| **C2F-SemiCD** | **80.93** | **81.05** | **80.80** | **90.64** | **74.72** | **67.96** | **82.61** | **79.04** | **86.53** | **91.05** | **76.60** | **70.38** |

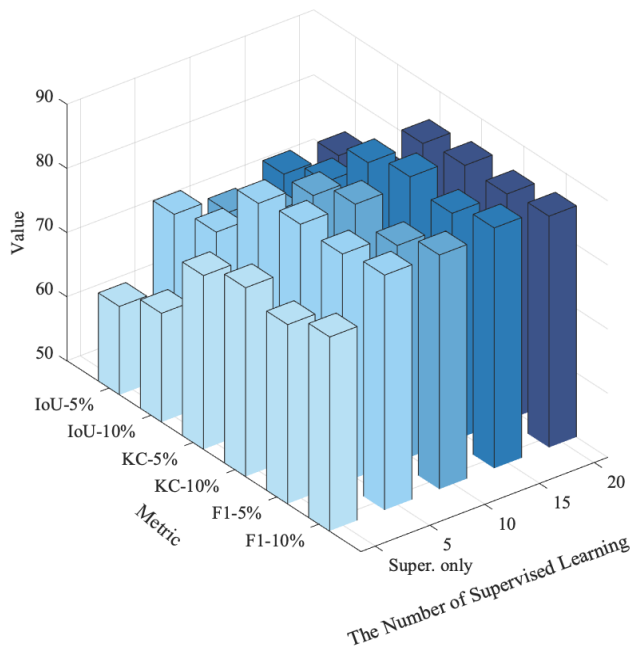

**Fig. 18.** Ablation experiments were performed on the WHU-CD data for the number of supervised learning at the beginning of the semi-supervised learning.

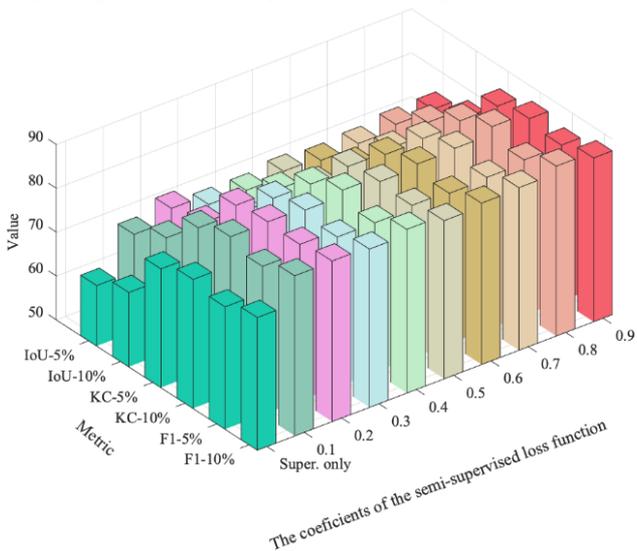

**Fig. 19.** Ablation experiments on the coefficients of the semi-supervised loss function in semi-supervised learning were performed on the WHU-CD data.

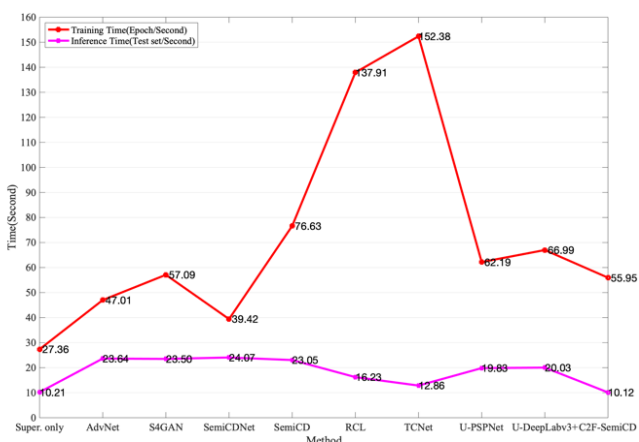

**Fig. 20.** Ablation experiments in comparing the training and inference times of the models on the Googlegz-CD dataset.

semi-supervised methods. The feature extraction ability of C2FNet exceeds other semi-supervised methods in some proportion of labelled samples, which directly proves the effectiveness of feature extraction, and it can still be improved on C2FNet in the future.